\documentclass{article}

\makeatletter
\def\input@path{{styles/}}
\makeatother
\usepackage[preprint]{colm2026_conference}
\usepackage{fontspec}

\renewcommand{\encodingdefault}{T1}
\normalfont
\usepackage{microtype}
\usepackage{graphicx}
\usepackage{trimclip}
\usepackage{xcolor}
\usepackage{booktabs}
\usepackage{array}
\usepackage{colortbl}
\usepackage{float}
\usepackage{tikz}
\usepackage{tcolorbox}
\usepackage{pgfplots}
\pgfplotsset{compat=1.18}
\usepgfplotslibrary{groupplots}
\usepackage{hyperref}
\usepackage{url}
\definecolor{abyss}{HTML}{121D36}
\definecolor{polarnight}{HTML}{1A2947}
\definecolor{nebula}{HTML}{2B3F66}
\definecolor{steeltrail}{HTML}{6D87BD}
\definecolor{skytrail}{HTML}{8FA8D8}
\definecolor{starlight}{HTML}{DFE7F5}
\definecolor{warmstar}{HTML}{E8D9C4}
\definecolor{allsparkwordmark}{HTML}{16233F}
\definecolor{allsparkspark}{HTML}{4A659C}
\definecolor{electricblue}{HTML}{3866FF}
\definecolor{covercream}{HTML}{EEF3FA}
\definecolor{coveraccent}{HTML}{3866FF}
\colorlet{pevekpurple}{skytrail}
\colorlet{bargray}{steeltrail}
\colorlet{barlgray}{starlight}

\newfontfamily\outfit{Outfit-Regular.ttf}[
  Path=./,
  BoldFont=Outfit-SemiBold.ttf
]

\hypersetup{
  colorlinks=true,
  linkcolor=electricblue,
  citecolor=electricblue,
  urlcolor=coveraccent,
  filecolor=electricblue
}
\setcitestyle{numbers,square,comma,sort&compress}

\usepackage{amsmath,amsfonts,bm}

\def\eqref#1{equation~\ref{#1}}

\def\1{\bm{1}}

\DeclareMathAlphabet{\mathsfit}{\encodingdefault}{\sfdefault}{m}{sl}
\SetMathAlphabet{\mathsfit}{bold}{\encodingdefault}{\sfdefault}{bx}{n}

\usepackage{multirow}
\usepackage{wrapfig}
\usepackage{caption}
\usepackage{framed}
\usepackage{longtable}

\newcommand{\reporttitle}{Vela: User Behavior as Supervision Sparks LLM Intelligence}
\newcommand{\reportauthors}{AllSpark Team}
\title{\reporttitle}
\author{\reportauthors}
\begin{document}

\fancyhead{}
\renewcommand{\headrulewidth}{0pt}
\color{abyss}
\thispagestyle{empty}

\vspace*{-0.44in}
\begin{tcolorbox}[
  width=\linewidth,
  colback=covercream,
  colframe=covercream,
  boxrule=0pt,
  arc=14pt,
  outer arc=14pt,
  boxsep=0pt,
  left=20pt,
  right=20pt,
  top=13pt,
  bottom=11pt
]
  {\outfit\fontsize{21.5}{25.5}\selectfont\bfseries\centering
    \textcolor{coveraccent}{Vela:} User Behavior as Supervision Sparks LLM Intelligence\par}
  \vspace{0.75em}
  {\bfseries\centering \reportauthors\par}

  \vspace{0.75em}
  \begingroup
  \normalfont
  \setlength{\parindent}{0pt}
  \setlength{\parskip}{0pt}
  Large language models are increasingly deployed in real-world applications,
from personal assistants that take actions based on users' long-term behavior histories
to recommendation and advertising systems that predict what users will do.
These
applications demand the user dimension of LLM intelligence: the
ability to infer a person's preferences, intent, and likely future
actions
from weeks to years of user behavior data. Yet current models
remain limited in this user dimension of
intelligence: they are trained
on static text or synthetic data, neither of which captures how user
behavior evolves over time. Real-world user behavior data, by contrast,
is already available at scale on content platforms and in personal-assistant agents.
These records capture sequences of user actions, making the user's actual
behavior a natural supervisory signal with the potential to improve LLM intelligence.
To this end, we propose
\emph{user behavior as supervision}, a paradigm in which the training
signal for LLMs comes from the user's real behavior in content platforms.
We further introduce \textbf{Vela}, a framework that operationalizes this paradigm
by converting real user behavior data into trainable tasks and trajectories. From the
anonymized behavior records of 100{,}000 users, Vela
generates 100{,}000 tasks across 91 real-world task templates.
Difficulty-based filtering and trajectory acceptance criteria jointly retain
approximately 75{,}000 tasks with selected training trajectories.
Fine-tuning Qwen3.6-35B-A3B on these training trajectories
improves performance over the base model by up to 11.5 points on user-related
tasks and by an average of 4 points on general benchmarks.
Scaling experiments further show that training on more user behavior data
improves user-related performance.
Reinforcement learning after fine-tuning outperforms reinforcement learning
from the base model by up to 7.7 points on user-related tasks.
These gains suggest that using real-world user behavior as supervision can improve not only user-centered capabilities but also
the broader intelligence of LLMs.

  \par
  \endgroup

  \vspace{0.65em}
  \noindent
  \begin{minipage}[c]{0.78\linewidth}
    \outfit\fontsize{8.4}{10.2}\selectfont
    \textbf{Date:} October 7, 2026\par
    \vspace{0.25em}
    \textbf{Code:} \href{https://github.com/AllSpark-Research/Vela}{https://github.com/AllSpark-Research/Vela}
  \end{minipage}\hfill
  \begin{minipage}[c]{0.20\linewidth}
    \raggedleft
    \includegraphics[height=16pt]{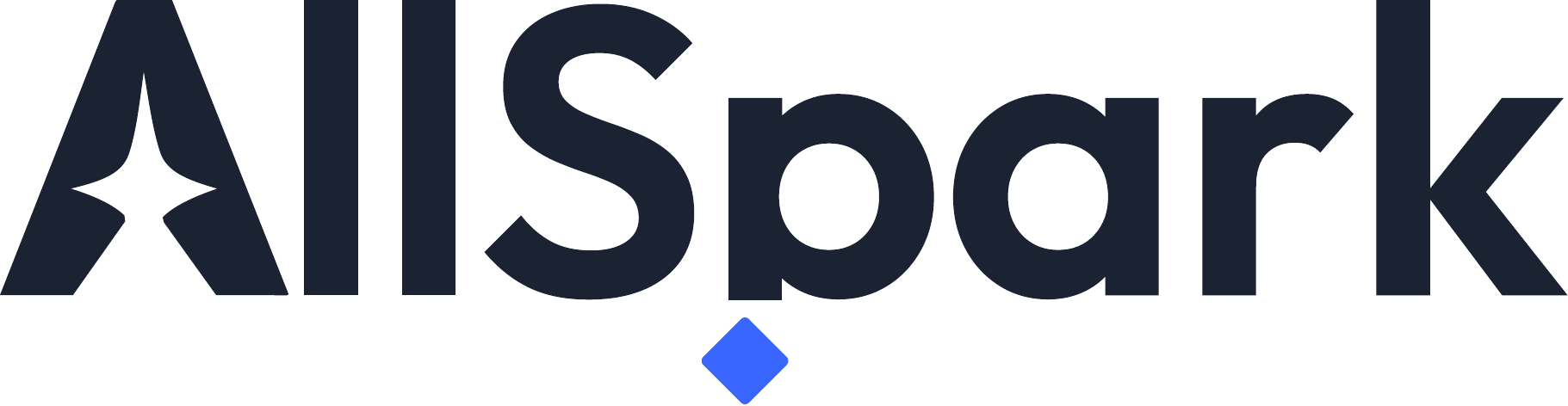}
  \end{minipage}
\end{tcolorbox}

\begin{center}
\begin{minipage}{\linewidth}
\centering
\includegraphics[width=\linewidth]{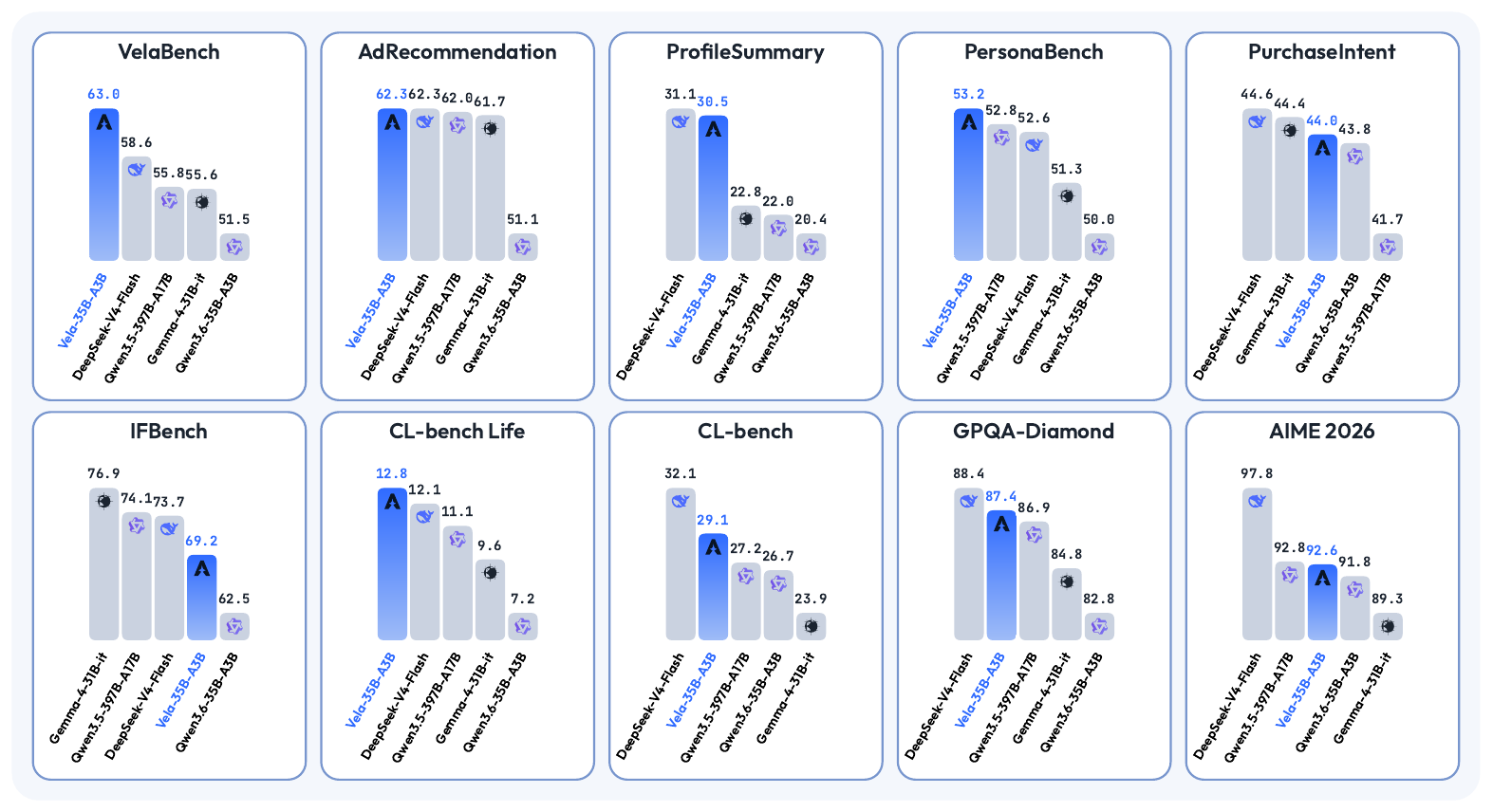}
\captionsetup{skip=4pt}
\captionof{figure}{\textbf{Vela-35B-A3B} (blue) against its base
model and three larger reference models across ten benchmarks. Within
each panel, bars are ordered by score with the highest on the left.}
\label{fig:teaser}
\end{minipage}
\end{center}

\clearpage
\raggedbottom
\begingroup
\setlength{\baselineskip}{1.06\baselineskip}
\section{Introduction}
\label{sec:intro}

As foundation models grow more capable, they are increasingly deployed
in real-world applications, from personal assistants that take actions
based on users' long-term behavior histories, such as OpenClaw \citep{openclaw2025} and
WorkBuddy\footnote{\url{https://www.codebuddy.cn/work/}}, to recommendation
and advertising systems that predict what users will do
\citep{coral2026,llmauction2025}.
Such scenarios share three properties:
(1) individualized: decisions concern one person, not the
average user;
(2) long-horizon: they span weeks or even years of
behavior;
(3) evidence-bound: every judgment must be grounded in the
user's behavior history.
Together, these properties define the user dimension of LLM
intelligence: the ability to infer an individual's preferences,
intent, and likely future actions from real-world user behavior data.
Understanding users from their behavior histories requires models to
learn from context, identify relevant signals amid noise, integrate
evidence over time, and track changes in preferences.

However, current LLMs remain limited in this user dimension of intelligence.
Common training sources include web pages, textbooks, code
\citep{su2025nemotroncc,stackv32026}, and synthetic data \citep{phi42024}.
These sources may contain user-related content, but such content may not capture how
individual users' behavior evolves over time.
Thus, training on these sources alone is insufficient to develop
the user dimension of LLM intelligence.
Recent evaluations show that LLMs with stronger general capabilities do not
always perform better in user-facing applications \citep{uxbench2026}.
LLMs also struggle to reliably complete tasks based on real user needs \citep{liveagent2026}.
Even when behavior histories are provided, models can drift toward an
average-user profile \citep{chen2026omnibehavior}, lose accuracy as histories
lengthen \citep{li2025finerob}, and struggle with personalization \citep{lunar2026}.

Real-world user behavior data, by contrast, is already available at scale
on content platforms and in personal-assistant agents
\citep{wang2026streamprofilebench,zhou2026openonerec,xu2025redrec}.
These records capture sequences of user actions and their context over weeks to years:
what the system presented, how the user responded, and how the user's
behavior evolved over time.
Given a user's behavior history, the actions the user actually takes afterward
provide a natural supervisory signal with the potential to improve LLM intelligence.
As public human-generated text becomes a potential bottleneck for LLM
scaling \citep{villalobos2024running}, user behavior data offers a large,
still-growing source of training signals.
The challenge is to convert heterogeneous behavior records into diverse
tasks with well-defined supervision.
Prior work has focused on converting user behavior data into training signals
for specific tasks.
For example, P5 unifies multiple recommendation tasks under a language
modeling objective \citep{geng2022p5}, while RLPF uses prediction feedback
to train LLMs to produce user summaries \citep{wu2024rlpf}.
However, these studies focus on specific applications, leaving open whether
training on user behavior data can improve performance across broader real-world
user-related scenarios and enhance general LLM capabilities.
This motivates three research questions:
\textbf{(1)} how to convert raw user behavior data into trainable
tasks and trajectories;
\textbf{(2)} to what extent user behavior data improves user-related
intelligence;
and \textbf{(3)} whether training on user behavior data strengthens or harms
general LLM capabilities.

To this end, we propose \emph{user behavior as supervision}, a paradigm
that trains LLMs to understand users from their behavior histories using
supervision derived from users' observed behavior.
We further introduce Vela, a framework that operationalizes this paradigm
by converting raw user behavior data from a large social media platform
into trainable tasks and trajectories.
Using anonymized behavior records from 100{,}000 users, Vela generates
100{,}000 tasks across 91 real-world task templates. Difficulty-based
filtering and trajectory acceptance criteria jointly retain approximately
75{,}000 tasks. The selected trajectories form two training corpora:
Vela-Instruct and Vela-RFT.
Fine-tuning Qwen3.6-35B-A3B on the training trajectories in Vela-Instruct and
Vela-RFT yields two models, Vela-35B-A3B and Vela-35B-A3B-RFT, respectively.
To evaluate user-related performance and general capabilities, we use
three groups of benchmarks:
\textbf{(1)} VelaBench: 982 held-out evaluation tasks drawn from the
same 91 task templates;
\textbf{(2)} user-related out-of-distribution (OOD) benchmarks: PurchaseIntent,
ProfileSummary, \mbox{PersonaBench}, AdRecommendation;
\textbf{(3)} general OOD benchmarks: AIME 2026,
GPQA-Diamond \citep{gpqa2023}, IFBench
\citep{ifbench2025}, CL-bench \citep{dou2026clbench}, CL-bench Life
\citep{dou2026clbenchlife}.
Vela-35B-A3B improves over the base model by 11.5 points on VelaBench
and by up to 11.2 points on user-related OOD benchmarks.
The average improvement on general OOD benchmarks is 4 points.
Scaling experiments further show that training on more user behavior data
improves user-related performance.
Reinforcement learning after fine-tuning outperforms reinforcement learning
from the base model by 7.7 points on both VelaBench and ProfileSummary.
These gains suggest that using real-world user behavior as supervision can improve not only user-centered capabilities but also
the broader intelligence of LLMs.

\par
\endgroup

\section{Vela: User Behavior as Supervision}
\label{sec:method}

\begin{figure}[t]
\centering
\captionsetup{skip=4pt}
\includegraphics[width=\linewidth]{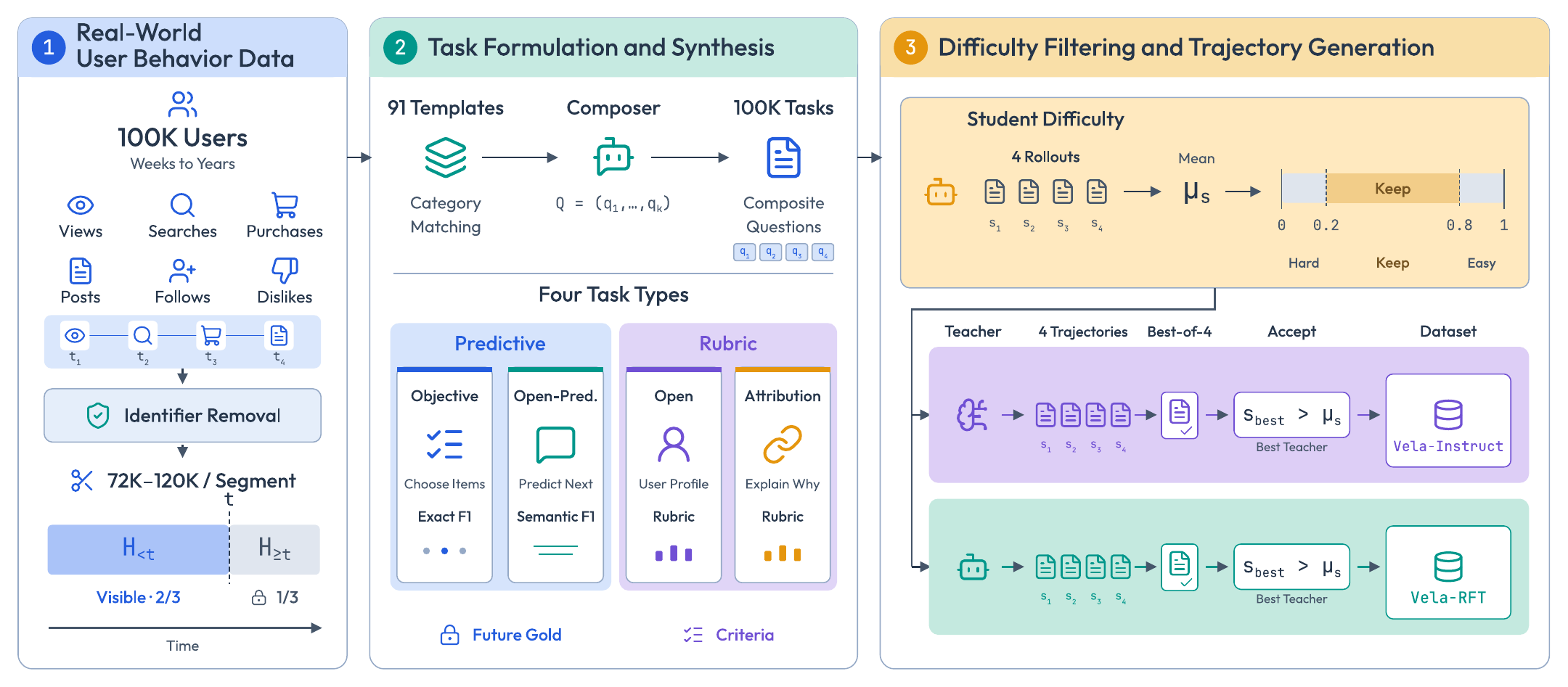}
\caption{Overview of Vela, a framework for user behavior as supervision. (1) Anonymized behavior records from
100{,}000 users are split into visible histories and hidden futures.
(2) A set of 91 question templates spanning two families, \emph{Predictive}
and \emph{Rubric}, is instantiated to construct 100{,}000 composite tasks.
Predictive questions are scored against ground-truth answers derived from
future behavior, while Rubric questions are scored using task-specific rubrics.
(3) Tasks are filtered by difficulty, and training trajectories are generated
and selected to form two training corpora: Vela-Instruct and Vela-RFT.}
\label{fig:framework}
\end{figure}

This section describes the Vela framework and evaluation setup.
We first describe the user behavior data (Section~\ref{sec:data}),
followed by task synthesis under user behavior as supervision
(Section~\ref{sec:synthesis}) and difficulty-based filtering and
trajectory generation (Section~\ref{sec:rollout}).
We then introduce the evaluation benchmarks (Section~\ref{sec:eval}).

\subsection{Real-world user behavior data}
\label{sec:data}
Real-world user behavior data consists of timestamped sequences of actions
taken by users on digital platforms.
This study is based on a large-scale corpus of anonymized user behavior
from a social media platform. The corpus contains behavior histories from
100{,}000 users, with each history spanning several weeks to several years.
Before task construction, we remove user profile information and direct
personal identifiers, including names, contact details, and user account IDs,
from the behavior records. Any behavior record containing personally
identifiable information is also excluded. The resulting anonymized histories
are used for task generation and model training.
We organize each user's records into six behavior categories:
(1) \textbf{Views}: browsing or watching content;
(2) \textbf{Searches}: searching for content or products;
(3) \textbf{Purchases}: purchasing products;
(4) \textbf{Posts}: creating and publishing content;
(5) \textbf{Follows}: following other accounts; and
(6) \textbf{Dislikes}: hiding posts to express negative feedback.
Illustrative examples of behavior records are provided in
Figure~\ref{fig:behavior-example} and Appendix~\ref{app:data-sources}.

\subsection{User behavior as supervision}
\label{sec:synthesis}
As illustrated in Figure~\ref{fig:framework}, under user behavior as supervision, each task combines multiple questions grounded in a single
user's behavior history. The questions span two
families: Predictive questions evaluated against observed future behavior,
and Rubric questions evaluated using task-specific criteria.

\textbf{Task formulation.} A user's behavior history is a sequence of
timestamped events, $H=(e_1,\dots,e_T)$, ordered chronologically.
Each event belongs to one of the six behavior categories defined in
Section~\ref{sec:data}. A cutoff index $t$ divides the sequence into a
\textbf{visible history}, $H_{<t}=(e_1,\dots,e_{t-1})$, and a
\textbf{hidden future}, $H_{\ge t}=(e_t,\dots,e_T)$.
A \textbf{task} $Q=(q_1,\dots,q_k)$ combines $k$ questions about the user.
Given the visible history and task questions, the model generates a response
$A=f(Q,H_{<t})$ containing an answer $a_i$ to each question $q_i$.
Each answer receives a score $s(a_i)\in[0,1]$, and the task score is the
mean across its $k$ questions, $\frac{1}{k}\sum_{i=1}^{k}s(a_i)$.

\begin{wrapfigure}{r}{0.5\linewidth}
\vspace{-\intextsep}
\centering
\includegraphics[width=\linewidth]{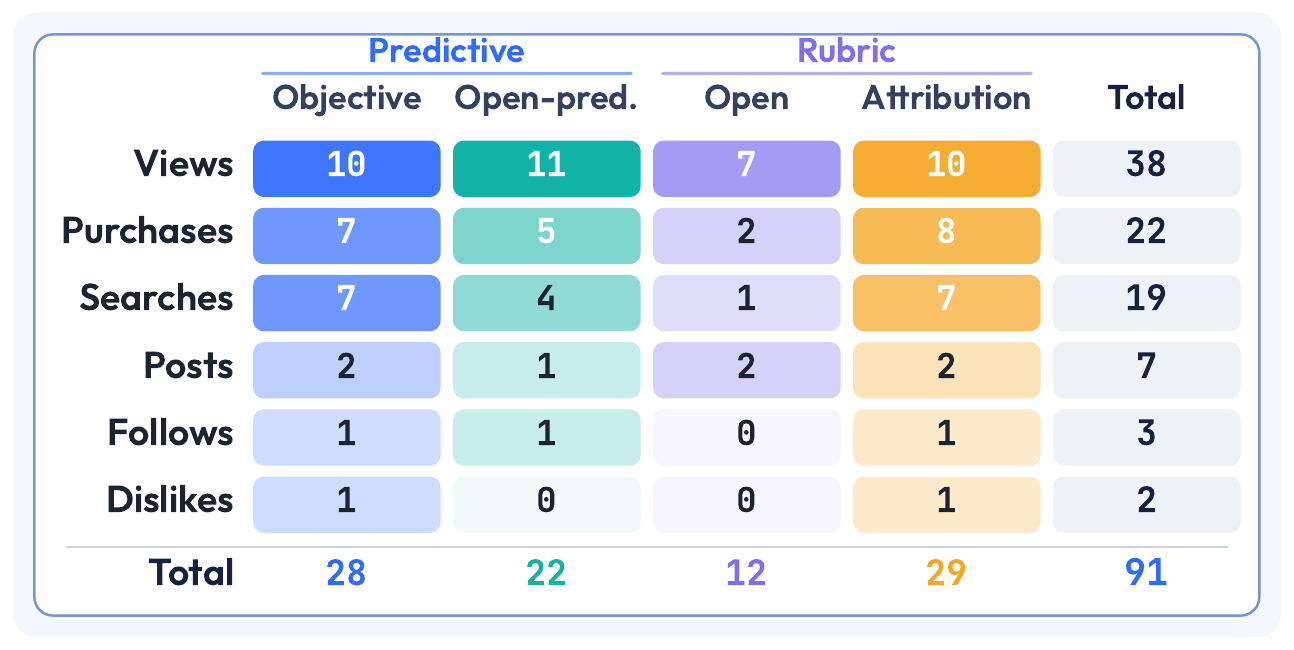}
\caption{Distribution of the 91 question templates across six behavior
categories and four question types. Objective and Open-prediction belong
to the Predictive family, while Open and Attribution belong to the Rubric
family. Cell values indicate template counts.}
\label{fig:templates}
\end{wrapfigure}

\textbf{Question templates.} We design 91 question templates across the six behavior categories,
covering four question types grouped into two families
(Figure~\ref{fig:templates}). \emph{Predictive} templates comprise Objective
and Open-prediction questions, whose reference answers are extracted
from the hidden future $H_{\ge t}$ using deterministic rules.
\emph{Rubric} templates comprise Open and Attribution questions, whose
answers are evaluated using task-specific criteria.
(1)~\textbf{Objective} questions provide the model with the visible
history $H_{<t}$ and a candidate list, and ask it to select the items or
entities involved in the user's behavior in the hidden future $H_{\ge t}$
(e.g.\ the products they will buy). The list contains one or more
ground-truth answers and distractors sampled from a pool of other users.
(2)~\textbf{Open-prediction} questions provide the model with the visible
history $H_{<t}$ and ask it to predict the user's behavior in the hidden
future $H_{\ge t}$ in free-form text, such as naming the products they
will buy, without a candidate list.
(3)~\textbf{Open} questions ask the model to summarize the user's
attributes, interests, or language style based on $H_{<t}$.
Because these summaries have no single correct answer, they are evaluated
against a task-specific rubric grounded in the visible history.
(4)~\textbf{Attribution} questions provide the model with the visible
history $H_{<t}$ and an observed user behavior from the hidden future
$H_{\ge t}$, and ask it to explain that behavior based on $H_{<t}$.
A task-specific rubric evaluates how well the explanation is supported
by the visible history.
We call this paradigm user behavior as supervision: observed
behavior provides the grounding for task construction and supervision.

\textbf{Metrics.} Predictive questions are evaluated by comparing the
LLM's predicted answers with ground-truth answers derived from the hidden future
$H_{\ge t}$. Let $Y_i$ denote the ground-truth answer set and $\hat{Y}_i$
the LLM's predicted answer set for question $q_i$.
For Objective questions, the selected candidates are matched exactly
against the ground-truth answers. For Open-prediction questions, an
LLM-based scorer receives the ground-truth answers and the evaluated LLM's
free-form response.
The scorer extracts the predicted items and maps semantically equivalent predictions
to their corresponding ground-truth items before computing set-F1:
\[
s(a_i)=\frac{2\,|\hat{Y}_i\cap Y_i|}{|\hat{Y}_i|+|Y_i|}.
\]
For Rubric questions, the scorer evaluates each answer against $m$
task-specific criteria, assigning a score $c_j\in[0,1]$ to each criterion.
The question score is the equal-weight mean of these scores,
$s(a_i)=\frac{1}{m}\sum_{j=1}^{m}c_j$.

\begin{figure}[t]
\centering
\captionsetup{skip=4pt}
\includegraphics[width=\linewidth]{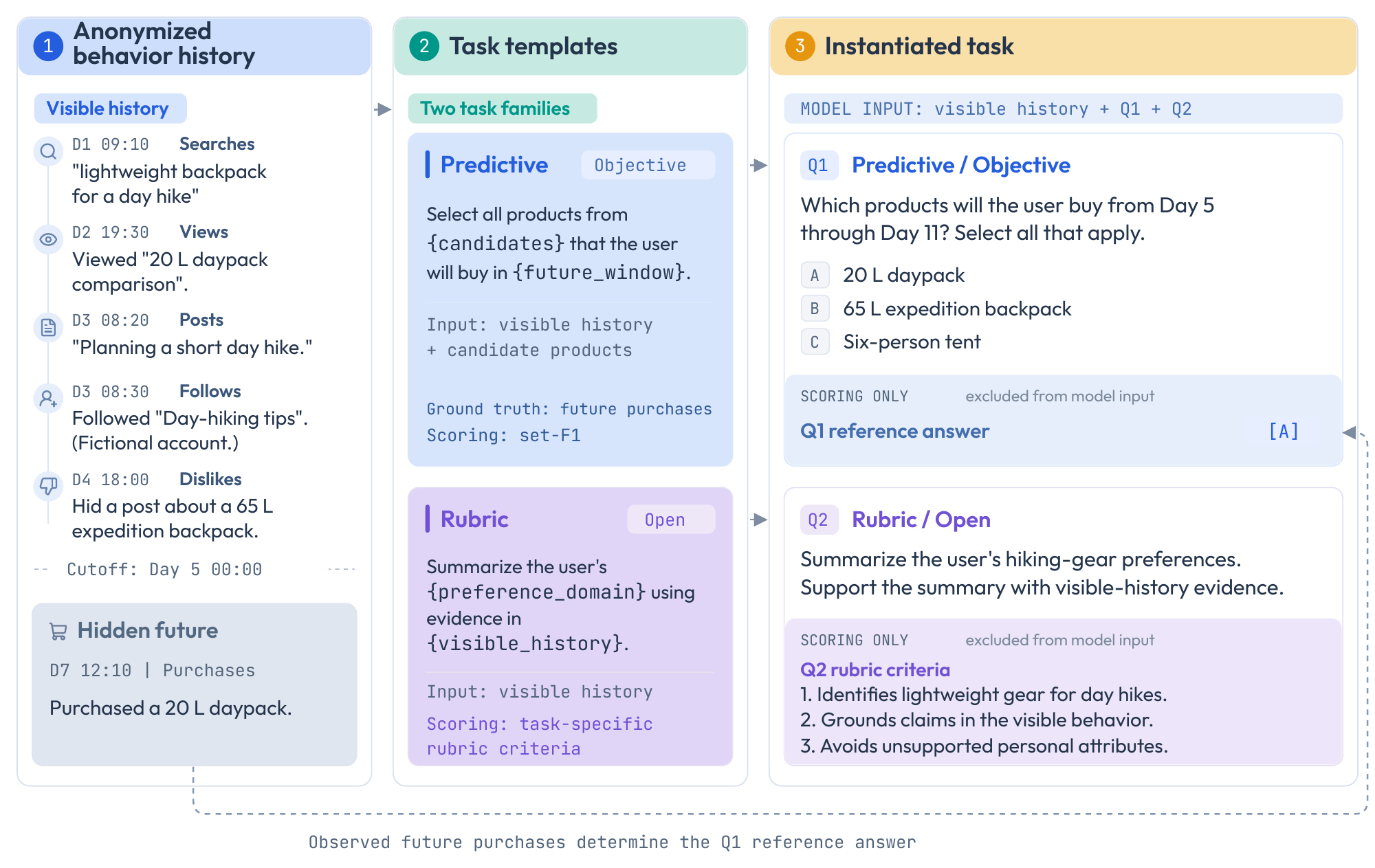}
\caption{An illustrative example of user behavior as supervision.
A behavior history and two templates form a task with Predictive and
Rubric questions. Future behavior provides the predictive ground truth,
while the scorer evaluates the Open response using a task-specific rubric.}
\label{fig:behavior-example}
\end{figure}

\textbf{Task synthesis.} We divide each user's behavior history into
segments with randomly sampled lengths between 72K and 120K tokens,
measured using the Qwen3.6-35B-A3B tokenizer. Each behavior record is kept
intact during segmentation. Within each segment, approximately the first
two-thirds of the tokens form the visible history $H_{<t}$, and the
remainder forms the hidden future $H_{\ge t}$. This division also
preserves complete behavior records.
For each segment, we select templates whose required behavior categories
are present. For the same future behavior event, we randomly select either a
Predictive template or an Attribution template. Including both would
reveal the prediction target through the Attribution question, allowing
the model to answer the Predictive question without inferring future
behavior from the visible history.
Predictive questions are constructed using predefined rules: Objective
questions include candidate lists, while Open-prediction questions require
free-form answers without candidates. For Open questions, an LLM-based
composer uses the selected template and visible history $H_{<t}$ to
generate a concrete question and its task-specific rubric. For Attribution
questions, the composer additionally uses an observed user behavior from
the hidden future $H_{\ge t}$ to construct a question that asks the model
to explain that behavior based on $H_{<t}$, together with a corresponding
rubric. The resulting questions are combined into one task $Q$ for each
segment.
Figure~\ref{fig:behavior-example} illustrates this process with an illustrative
example. In total, this yields 100{,}000 candidate tasks from 100{,}000 users.
Both the composer and scorer use Qwen3.8-2.4T-A95B with role-specific prompts
(Appendix~\ref{app:prompts}). Both use the xhigh thinking level.

\subsection{Difficulty-based filtering and trajectory generation}
\label{sec:rollout}
As illustrated in part (3) of Figure~\ref{fig:framework}, training
trajectories are constructed in two stages: candidate tasks are first
filtered by difficulty, and trajectories are then generated and selected
for the retained tasks.

\textbf{Difficulty-based filtering.} Qwen3.6-35B-A3B generates four
responses to each task with thinking disabled. The scorer evaluates each
response, and the mean score is used to estimate task difficulty.
Tasks with mean scores in $[0.2,0.8]$ are retained, focusing training on
tasks of intermediate difficulty.

\textbf{Trajectory generation.} For each retained task, each generation
model produces four training trajectories. Only the highest-scoring
trajectory is considered for inclusion. This trajectory is accepted only
if its score exceeds the mean score obtained during difficulty filtering;
otherwise, no trajectory from that model is retained for the task.
We use Qwen3.8-2.4T-A95B at the xhigh thinking level and Qwen3.6-35B-A3B
with thinking enabled to generate Vela-Instruct and Vela-RFT, respectively.
Vela-RFT contains trajectories generated by Qwen3.6-35B-A3B itself.
For Vela-Instruct, difficulty-based filtering and the trajectory acceptance
criterion jointly retain approximately 75{,}000 of the 100{,}000 candidate
tasks, with one training trajectory per retained task. Vela-RFT is constructed
on a smaller scale and contains 9{,}375 training samples.

\subsection{Evaluation benchmarks}
\label{sec:eval}
To evaluate user-related performance and general capabilities, we use
three groups of benchmarks: VelaBench for in-domain performance,
user-related OOD benchmarks for performance in real-world user-facing
scenarios, and general OOD benchmarks for reasoning,
instruction following, and context learning.

\textbf{VelaBench.} VelaBench evaluates in-domain performance using
982 held-out tasks constructed with the 91 question templates,
covering all four question types and six behavior
categories. Users included in VelaBench are excluded from training-data
construction.
Each task is scored using the protocol in Section~\ref{sec:synthesis}:
set-F1 for Objective and Open-prediction questions, and the mean rubric
score for Open and Attribution questions.

\textbf{User-related OOD.} We use four independently constructed
benchmarks based on real user behavior to evaluate LLM capabilities in
real-world user-facing scenarios. These benchmarks cover purchase intent
prediction, profile summarization, persona understanding, and ad recommendation.
Their inputs, outputs, and metrics are described below, with examples in
Appendix~\ref{app:benches}:
(1) \textbf{PurchaseIntent}: given a user's history and a set of
candidate items, rank the candidate items by how likely the user is to
purchase them, evaluated by recall@20 against
ground-truth purchases;
(2) \textbf{ProfileSummary}: given a user's history, generate a profile and
interest summary in a structured format, evaluated using rubrics for the user profile, core
interests, and inferred interests;
(3) \textbf{PersonaBench}: given a user's history, answer a
multiple-choice question about the user's attributes or purchase behavior, scored by
macro-F1;
(4) \textbf{AdRecommendation}: given a user's profile and recent search
queries, generate search queries to retrieve ads relevant to the user.
An LLM-based scorer evaluates the generated queries against rubrics for inferred intent and subsequent
ad interactions.

\textbf{General OOD.} We evaluate general capabilities using five
benchmarks: AIME 2026 \citep{aime2026} for mathematics, GPQA-Diamond
\citep{gpqa2023} for science, IFBench \citep{ifbench2025} for
instruction following, and CL-bench \citep{dou2026clbench} and CL-bench
Life \citep{dou2026clbenchlife} for context learning.

Among these benchmarks, PurchaseIntent, PersonaBench, AIME 2026,
GPQA-Diamond, and IFBench use
rule-based evaluation.
Outputs requiring semantic matching or rubric-based evaluation in
VelaBench, ProfileSummary, AdRecommendation, CL-bench, and CL-bench Life
are assessed by the scorer, Qwen3.8-2.4T-A95B, at the xhigh thinking level.
The evaluated models use thinking mode, with the highest supported
thinking level where configurable.

\section{Experiments}
\label{sec:exp}


\definecolor{ebl}{HTML}{2f6bff}
\definecolor{tel}{HTML}{14B8A6}
\newcommand{\vsep}{\color{black!35}\vrule width 0.7pt}
\newcommand{\gp}[1]{\,\textcolor{red}{\scriptsize($+$#1)}}
\newcommand{\gm}[1]{\,\textcolor{green!45!black}{\scriptsize($-$#1)}}
\newcommand{\vch}[1]{\begin{tabular}[c]{@{}c@{}}#1\end{tabular}}
\newlength{\velascorewidth}
\newlength{\veladeltawidth}

\begin{table}[t]
\centering
\small
\setlength{\tabcolsep}{4.5pt}
\newcommand{\velascore}[2]{%
  \makebox[\velascorewidth][r]{#1}\makebox[\veladeltawidth][r]{#2}}
\settowidth{\velascorewidth}{00.0}
\settowidth{\veladeltawidth}{\gp{11.5}}
\caption{Performance of Vela-35B-A3B, Vela-35B-A3B-RFT, and the baselines
across ten benchmarks. Results are grouped into \textbf{In-domain} (VelaBench),
\textbf{User-related OOD}, and \textbf{General OOD}.
\textbf{Avg.} is the mean within each OOD group.
Parentheses give the change
over the base model (Qwen3.6-35B-A3B): {\color{red}red} for
gains, {\color{green!45!black}green} for drops.}
\label{tab:main}
\resizebox{\linewidth}{!}{%
\begin{tabular}{l ccccc >{\columncolor{ebl!8}}c >{\columncolor{ebl!8}}c >{\columncolor{ebl!22}}c}
\toprule
Benchmark & \vch{Gemma-4-\\31B-it} & \vch{Qwen3.5-\\397B-A17B} & \vch{DeepSeek-\\V4-Flash} & \vch{Qwen3.8-\\2.4T-A95B} & \vch{Kimi-K3} & \vch{Qwen3.6-\\35B-A3B} & \vch{Vela-35B-\\A3B-RFT} & \vch{Vela-35B-\\A3B} \\
\midrule
\textit{Size} & 31B & 397B & 284B & 2.4T & 2.8T & 35B & 35B & 35B \\
\midrule
\multicolumn{5}{l}{\textit{In-domain}} & \multicolumn{4}{c}{}\\
VelaBench & 55.6 & 55.8 & 58.6 & 62.3 & 64.0 & 51.5 & \velascore{53.5}{\gp{2.0}} & \velascore{63.0}{\gp{11.5}} \\
\midrule
\multicolumn{5}{l}{\textit{User-related OOD}} & \multicolumn{4}{c}{}\\
PurchaseIntent & 44.4 & 41.7 & 44.6 & 45.5 & 44.9 & 43.8 & \velascore{42.9}{\gm{0.9}} & \velascore{44.0}{\gp{0.2}} \\
ProfileSummary & 22.8 & 22.0 & 31.1 & 34.1 & 38.3 & 20.4 & \velascore{23.0}{\gp{2.6}} & \velascore{30.5}{\gp{10.1}} \\
PersonaBench & 51.3 & 52.8 & 52.6 & 55.9 & 54.6 & 50.0 & \velascore{50.3}{\gp{0.3}} & \velascore{53.2}{\gp{3.2}} \\
AdRecommendation & 61.7 & 62.0 & 62.3 & 62.5 & 65.0 & 51.1 & \velascore{55.5}{\gp{4.4}} & \velascore{62.3}{\gp{11.2}} \\
\addlinespace[3pt]
\textit{Avg.} & 45.1 & 44.6 & 47.7 & 49.5 & 50.7 & 41.3 & \velascore{42.9}{\gp{1.6}} & \velascore{47.5}{\gp{6.2}} \\
\midrule
\multicolumn{5}{l}{\textit{General OOD}} & \multicolumn{4}{c}{}\\
AIME 2026 & 89.3 & 92.8 & 97.8 & 92.5 & 91.0 & 91.8 & \velascore{92.3}{\gp{0.5}} & \velascore{92.6}{\gp{0.8}} \\
GPQA-Diamond & 84.9 & 86.9 & 88.4 & 87.9 & 91.9 & 82.8 & \velascore{80.8}{\gm{2.0}} & \velascore{87.4}{\gp{4.6}} \\
IFBench & 76.9 & 74.1 & 73.7 & 79.4 & 72.1 & 62.5 & \velascore{64.2}{\gp{1.7}} & \velascore{69.2}{\gp{6.7}} \\
CL-bench & 23.9 & 27.2 & 32.1 & 35.2 & 41.0 & 26.7 & \velascore{26.0}{\gm{0.7}} & \velascore{29.1}{\gp{2.4}} \\
CL-bench Life & 9.6 & 11.1 & 12.1 & 13.3 & 16.8 & 7.2 & \velascore{7.9}{\gp{0.7}} & \velascore{12.8}{\gp{5.6}} \\
\addlinespace[3pt]
\textit{Avg.} & 56.9 & 58.4 & 60.8 & 61.7 & 62.6 & 54.2 & \velascore{54.2}{\,\textcolor{black!45}{\scriptsize(0.0)}} & \velascore{58.2}{\gp{4.0}} \\
\bottomrule
\end{tabular}%
}
\end{table}

\subsection{Setup}
We fine-tune Qwen3.6-35B-A3B \citep{qwen3635ba3b} on Vela-Instruct and Vela-RFT to obtain
\textbf{Vela-35B-A3B} and \textbf{Vela-35B-A3B-RFT}, respectively.
Both models are trained using the Swift framework \citep{zhao2024swift}
with the same training settings: a packed sequence length of 131{,}072
tokens, two epochs, and a learning rate of $1\times10^{-5}$.
Both models are evaluated on the three benchmark groups described in
Section~\ref{sec:eval}.
We compare against the base model, Qwen3.6-35B-A3B, and five additional
models: Gemma-4-31B-it \citep{gemmateam2026gemma4},
Qwen3.5-397B-A17B \citep{qwen35397ba17b}, DeepSeek-V4-Flash
\citep{deepseekai2026v4}\footnote{Throughout this paper, DeepSeek-V4-Flash refers to the
DeepSeek-V4-Flash-0731 checkpoint.}, Qwen3.8-2.4T-A95B \citep{qwenlm2026qwen38},
and Kimi-K3 \citep{moonshotai2026kimik3}. To investigate whether scaling
user behavior data improves LLM intelligence,
we also train on five random subsets of Vela-Instruct, at 4.7K,
9.4K, 18.8K, 37.5K, and 75K examples. Unless otherwise stated,
Vela-35B-A3B refers to the model trained at the 75K scale throughout
this paper.

\begin{figure}[t]
\centering
\captionsetup{skip=4pt}
\includegraphics[width=\linewidth]{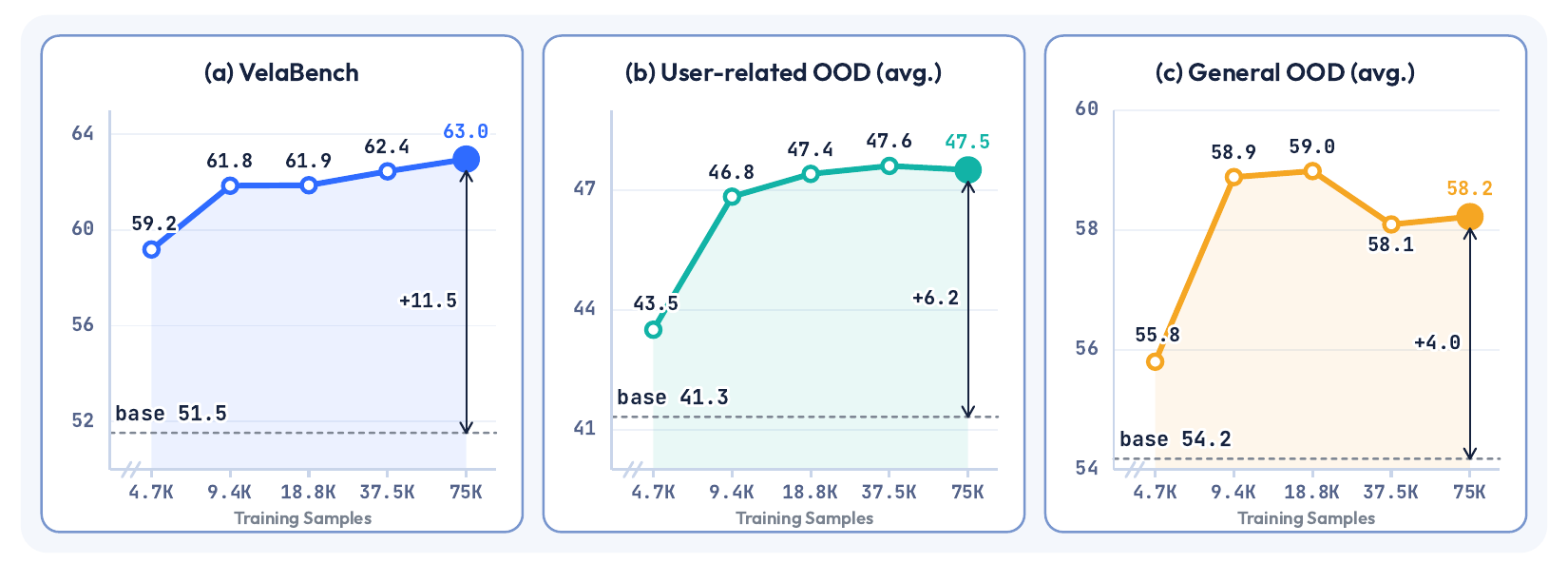}
\caption{Performance of Qwen3.6-35B-A3B fine-tuned on Vela-Instruct
at different data scales: (a) VelaBench, (b) average user-related OOD
performance, and (c) average general OOD performance. Dashed lines
indicate base model performance.}
\label{fig:scaling}
\end{figure}

\subsection{User behavior as supervision improves user-related intelligence}

Table~\ref{tab:main} compares Vela-35B-A3B and Vela-35B-A3B-RFT with
the baseline models across ten benchmarks. Vela-35B-A3B outperforms
Qwen3.6-35B-A3B on all five user-related benchmarks.
On VelaBench, Vela-35B-A3B scores 63.0, exceeding the base model by
11.5 points. With only 35B total parameters, Vela-35B-A3B outperforms
larger DeepSeek-V4-Flash and Qwen3.8-2.4T-A95B on this benchmark.
Across the four user-related OOD benchmarks, Vela-35B-A3B improves
over the base model by 6.2 points on average, with the largest gain
of 11.2 points on AdRecommendation.
Vela-35B-A3B-RFT also improves over the base model, by 2.0 points on
VelaBench and 1.6 points on average across the user-related OOD benchmarks.
These results suggest that user behavior as supervision improves
user-related performance both on VelaBench and in real-world
user-facing scenarios.

\subsection{User behavior as supervision benefits general LLM intelligence}
\label{sec:general}
As shown in Table~\ref{tab:main}, Vela-35B-A3B outperforms
Qwen3.6-35B-A3B on all five general OOD benchmarks, with an average
gain of 4.0 points. The largest gains are on IFBench and CL-bench Life,
with improvements of 6.7 and 5.6 points, respectively.
On CL-bench and CL-bench Life, Vela-35B-A3B also outperforms
Qwen3.5-397B-A17B by 1.9 and 1.7 points, respectively.
With training trajectories generated by Qwen3.6-35B-A3B itself,
Vela-35B-A3B-RFT improves user-related performance while maintaining
the base model's average score on general OOD benchmarks.
Together, these results show that the benefits of user behavior as supervision extend beyond user-related tasks to general LLM capabilities.

\subsection{Scaling user behavior data improves LLM intelligence}
\label{sec:scaling}

Figure~\ref{fig:scaling} reports scores of Qwen3.6-35B-A3B fine-tuned
on five random subsets of Vela-Instruct, from 4.7K to 75K samples.
On VelaBench, 4.7K training examples already raise the score from
51.5 to 59.2. Increasing the training set to 75K examples further
raises the score to 63.0. VelaBench performance continues to improve
from 37.5K to 75K examples, suggesting potential benefits from further scaling.
The user-related OOD average rises from 43.5 at 4.7K to 47.5 at 75K.
The general OOD average also increases, from 55.8 at 4.7K to 58.2 at
75K, with all tested scales outperforming the base model.
Together, these results demonstrate that scaling user behavior data
from 4.7K to 75K examples improves both user-related and general LLM capabilities.

\section{Analysis}
\label{sec:analysis}

\subsection{User behavior as supervision provides a stronger foundation for downstream tasks}
\label{sec:rl}
\begin{figure}[t]
\centering
\captionsetup{skip=4pt}
\includegraphics[width=\linewidth]{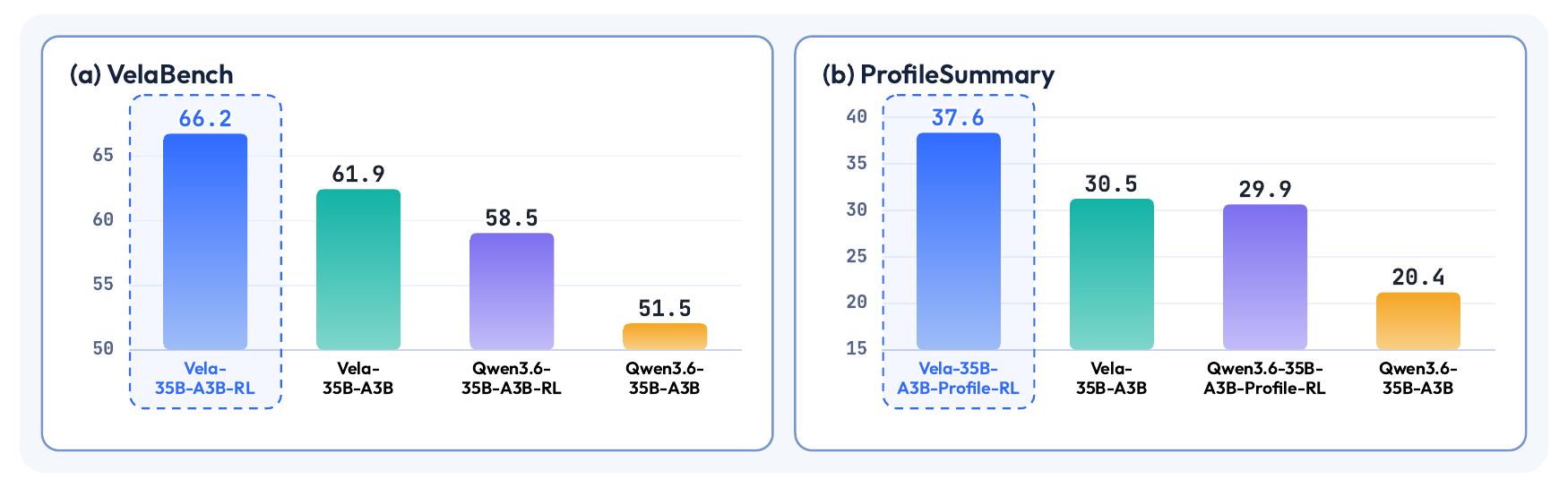}
\caption{Performance before and after GSPO on (a) VelaBench and
(b) ProfileSummary, starting from Qwen3.6-35B-A3B or Vela-35B-A3B.
Vela-35B-A3B here denotes the SFT checkpoint trained on 18.8K
Vela-Instruct examples.}
\label{fig:rl}
\end{figure}

We further investigate whether user behavior as supervision provides
a stronger foundation for downstream tasks.
Reinforcement learning is commonly applied after supervised fine-tuning
in post-training pipelines. We therefore investigate whether the
stronger foundation provided by user behavior as supervision leads to
better performance after reinforcement learning.
We apply GSPO \citep{gspo2025} with the same
configuration (Appendix~\ref{app:rl-config})
to Qwen3.6-35B-A3B and Vela-35B-A3B.
For the VelaBench experiment, we randomly sample 2{,}000 tasks from
the training set for RL. For the ProfileSummary experiment, we use
2{,}330 training examples from the same task domain as ProfileSummary.
For these experiments, Vela-35B-A3B denotes the checkpoint trained on
18.8K Vela-Instruct examples. We select this checkpoint because it
achieves the highest overall score across all ten benchmarks among
the data scales evaluated in Section~\ref{sec:scaling}.
Each resulting model is evaluated on the benchmark corresponding to
its RL training domain.

Figure~\ref{fig:rl} reports the results. On VelaBench, RL initialized
from Vela-35B-A3B reaches 66.2, compared with 58.5 when initialized
from Qwen3.6-35B-A3B. On ProfileSummary, the corresponding scores
are 37.6 and 29.9. With the same RL configuration, initialization
from Vela-35B-A3B yields scores 7.7 points higher on both benchmarks.
On both benchmarks, RL from the base model remains below the SFT
checkpoint before RL, while RL from the SFT checkpoint further
improves performance.
These results show that user behavior as supervision provides a
stronger foundation for reinforcement learning, leading to better
final performance under the same RL configuration.

\subsection{Difficulty-based filtering improves user-related intelligence}
\label{sec:ablation}
To investigate the contribution of difficulty-based filtering
(Section~\ref{sec:rollout}), we compare models fine-tuned on training
examples constructed with and without this filtering step.
We conduct the ablation at half the full training scale, using 37.5K
examples and the corresponding checkpoint from Section~\ref{sec:scaling}
as the filtered baseline.
The variant without filtering draws tasks randomly from the 100{,}000
candidates without applying the difficulty threshold.
Both variants use the same trajectory generation procedure,
trajectory acceptance criterion, and fine-tuning settings, with
37.5K final training examples each.
As shown in Table~\ref{tab:ablation-screen}, difficulty-based filtering
improves performance on all four user-related OOD benchmarks, with
an average gain of 1.1 points, while VelaBench performance remains
unchanged at 62.4.
These results show that difficulty-based filtering improves user-related
OOD performance while preserving in-domain performance.

\begin{table}[t]
\centering
\small
\caption{Ablation of difficulty-based filtering
(Section~\ref{sec:rollout}). Both models are fine-tuned on 37.5K
training examples. $\Delta$ denotes the score difference between
models trained with and without difficulty-based filtering.}
\label{tab:ablation-screen}
\resizebox{\linewidth}{!}{%
\begin{tabular}{l c cccc}
\toprule
 & In-dom. & \multicolumn{4}{c}{\textit{User-related OOD}} \\
\cmidrule(lr){3-6}
& VelaBench & PurchaseIntent & ProfileSummary & PersonaBench & AdRecommendation \\
\midrule
Qwen3.6-35B-A3B & 51.5 & 43.8 & 20.4 & 50.0 & 51.1 \\
w/o difficulty-based filtering & 62.4 & 44.6 & 29.0 & 51.5 & 61.0 \\
w/ difficulty-based filtering & 62.4 & 45.4 & 30.5 & 52.7 & 61.9 \\
\midrule
$\Delta$ & \textcolor{black!45}{\scriptsize($+$0.0)} & \gp{0.8} & \gp{1.5} & \gp{1.2} & \gp{0.9} \\
\bottomrule
\end{tabular}%
}
\end{table}

\section{Related Work}
\label{sec:related}

Recent work evaluates how well LLMs remember, understand, and respond to
individual users over time. LongMemEval \citep{wu2024longmemeval} examines
information retention and reasoning across long conversational histories,
while PrefEval \citep{hou2025prefeval} tests whether models infer and follow
explicit and implicit user preferences. PersonaMem \citep{song2025personamem}
focuses on evolving user profiles and personalized responses, and Memora
\citep{memora2026} examines how agents update memory and avoid relying on
outdated information. ATM-Bench \citep{mei2026atmbench} extends personalized
memory evaluation to multiple modalities and sources, including questions
that require resolving personal references and combining evidence.
Behavior prediction benchmarks further connect user understanding to
observed behavior: OPeRA \citep{wang2025opera} captures online shopping
sessions with observations, personas, rationales, and actions; FineRob
\citep{li2025finerob} evaluates fine-grained social media behavior prediction
from accumulated histories; and OmniBehavior \citep{chen2026omnibehavior}
studies long-horizon behavior across scenarios. Our work complements
these efforts by using real-world behavior histories as supervision
for a broad set of user-related tasks.

Learning from user histories has also been explored through user
representations, recommendation, and behavior simulation. User-LLM
\citep{userllm2025} encodes interaction histories into user embeddings that
condition LLM responses, while P5 \citep{geng2022p5} expresses multiple
recommendation tasks within a unified text-to-text framework. RLPF
\citep{wu2024rlpf} uses prediction feedback to optimize user summaries for
downstream tasks. DEEPER \citep{deeper2025} similarly uses
discrepancies between predicted and observed behavior to refine dynamic
personas. Beyond user representations, Customer-R1 \citep{customerR1}
uses reinforcement learning to improve personalized shopping behavior
simulation, and the OM-CoT framework accompanying FineRob
\citep{li2025finerob} fine-tunes models for behavior prediction with
long-term memory. The source of training signals also matters: work on
recursive training with model-generated data identifies risks of losing
information from the original data distribution \citep{shumailov2024ai}.
Vela combines tasks and trajectories with supervision
grounded in real user behavior, using observed future behavior to derive
predictive ground truth. We study this supervision across a broad task
collection and multiple data scales, evaluating its effects on user-related
performance, general LLM capabilities, and subsequent reinforcement learning.

\section{Conclusion}
User behavior as supervision derives training signals from users' observed
behavior, and Vela turns real-world behavior histories into
training tasks and trajectories. Experiments demonstrate gains in both
user-related and general LLM capabilities, further improvements with
data scaling, and a stronger foundation for downstream tasks through
reinforcement learning. These findings establish real-world user behavior
as a scalable source of supervision and motivate extending this approach
to more real-world tasks and scenarios.

\section*{Full Author List}
\begin{minipage}{\linewidth}
Yang Xiao\textsuperscript{\textcolor{electricblue}{*}},
Yusong Sun\textsuperscript{\textcolor{electricblue}{*}},
Haoyi Wu, Wenyang Hui, Weiqiang Hang, Luqing Wang, Zejian Wang,
Zirui Tu, Jingsheng Mao, Shaobo Hu, Yang Luo, Jiale Guo, Jin Zhang,
Yudan Liu, Zhaokai Luo, Ren Huan, Yao Hu,
Chengyue Jiang\textsuperscript{\textcolor{electricblue}{\dag}},
Wenjie Li, Chuan Mu
\par\medskip
\noindent\rule{0.32\linewidth}{0.4pt}\par
{\small
\textsuperscript{*}These authors contributed equally to this work.\par
\textsuperscript{\dag}Project lead.\par
Contact: \href{mailto:yang-alan.xiao@connect.polyu.hk}{\nolinkurl{yang-alan.xiao@connect.polyu.hk}}\par}
\end{minipage}

\subsection*{AI use statement}
Generative AI tools assisted code development, literature retrieval,
interpretation of experimental results, and manuscript drafting and
revision. LLMs also served as task composers,
trajectory generators, and scorers in the research workflow described in
Section~\ref{sec:method}. The authors reviewed AI-assisted outputs and
retain responsibility for the methods, results, and claims in this paper.

\subsection*{Ethics statement}
The use of behavior data in this study was authorized by the platform
and underwent privacy and ethics review. Before task construction,
user profile information and direct personal identifiers were removed,
and records containing personally identifiable information were excluded
(Section~\ref{sec:data}).
These measures are intended to reduce privacy risks when studying
user understanding from behavioral histories.

\subsection*{Reproducibility statement}
We will publicly release task templates, generation and scoring prompts,
and training and evaluation code.
Section~\ref{sec:method} describes the data processing, task construction,
difficulty-based filtering, trajectory selection, and scoring procedures.
Section~\ref{sec:exp} specifies the fine-tuning settings, baseline models,
and data scales, while Sections~\ref{sec:rl} and~\ref{sec:ablation}
describe the reinforcement learning comparisons and filtering ablation.
Illustrative behavior records and representative task templates are provided in
Appendices~\ref{app:records} and~\ref{app:catalog} to clarify the input
format, question construction, and supervision used by Vela.

\bibliography{bibliography/references}

\begin{thebibliography}{40}
\providecommand{\natexlab}[1]{#1}
\providecommand{\url}[1]{\texttt{#1}}
\expandafter\ifx\csname urlstyle\endcsname\relax
  \providecommand{\doi}[1]{doi: #1}\else
  \providecommand{\doi}{doi: \begingroup \urlstyle{rm}\Url}\fi

\bibitem[Abdin et~al.(2024)Abdin, Aneja, Behl, Bubeck, Eldan, Gunasekar,
  Harrison, Hewett, Javaheripi, Kauffmann, Lee, Lee, Li, Liu, Mendes, Nguyen,
  Price, de~Rosa, Saarikivi, Salim, Shah, Wang, Ward, Wu, Yu, Zhang, and
  Zhang]{phi42024}
Marah Abdin, Jyoti Aneja, Harkirat Behl, Sébastien Bubeck, Ronen Eldan, Suriya
  Gunasekar, Michael Harrison, Russell~J. Hewett, Mojan Javaheripi, Piero
  Kauffmann, James~R. Lee, Yin~Tat Lee, Yuanzhi Li, Weishung Liu, Caio C.~T.
  Mendes, Anh Nguyen, Eric Price, Gustavo de~Rosa, Olli Saarikivi, Adil Salim,
  Shital Shah, Xin Wang, Rachel Ward, Yue Wu, Dingli Yu, Cyril Zhang, and
  Yi~Zhang.
\newblock {Phi-4} technical report, 2024.
\newblock URL \url{https://arxiv.org/abs/2412.08905}.

\bibitem[Azhar et~al.(2026)Azhar, Zhou, Jiang, Wang, Sharma, DeSousa, Liu, Guo,
  Zhang, and Fan]{coral2026}
Muhammad~Rafay Azhar, Yuhang Zhou, Gilbert Jiang, Yuchen Wang, Rahul Sharma,
  Matthew DeSousa, Jiayi Liu, Xin Guo, Lizhu Zhang, and Xiangjun Fan.
\newblock {CORAL}: An {LLM}-native harness for production recommender systems,
  2026.
\newblock URL \url{https://arxiv.org/abs/2609.02730}.

\bibitem[Chen et~al.(2025)Chen, Du, Chen, Xu, Zhang, Yuan, Chen, Li, and
  Xiao]{deeper2025}
Aili Chen, Chengyu Du, Jiangjie Chen, Jinghan Xu, Yikai Zhang, Siyu Yuan,
  Zulong Chen, Liangyue Li, and Yanghua Xiao.
\newblock {DEEPER} insight into your user: Directed persona refinement for
  dynamic persona modeling, 2025.
\newblock URL \url{https://arxiv.org/abs/2502.11078}.

\bibitem[Chen et~al.(2026)Chen, Xu, Cao, Pan, Zhang, Hu, Du, Gao, Lu, Sun, Han,
  Sun, Wu, and Lin]{chen2026omnibehavior}
Jiawei Chen, Ruoxi Xu, Boxi Cao, Ruotong Pan, Yunfei Zhang, Yifei Hu, Yong Du,
  Tingting Gao, Yaojie Lu, Yingfei Sun, Xianpei Han, Le~Sun, Xiangyu Wu, and
  Hongyu Lin.
\newblock Towards real-world human behavior simulation: Benchmarking large
  language models on long-horizon, cross-scenario, heterogeneous behavior
  traces, 2026.
\newblock URL \url{https://arxiv.org/abs/2604.08362}.

\bibitem[DeepSeek-AI et~al.(2026)DeepSeek-AI, Xu, Lin, Xue, Wang, Xu, Wu,
  Zhang, Lin, Dong, Ling, Lu, Zhao, Deng, Hou, Xu, Shao, Ruan, Sun, Dai, Guo,
  Yang, Chen, Li, Ji, Li, Wei, Lin, Yuan, Xia, Dai, Hao, Chen, Cao, Meng, Li,
  Yu, Zhang, Xu, Li, Liang, Zhang, Luo, Wei, Yuan, Zhang, Luo, Chen, Ji, Zhang,
  Ding, Tang, Cao, Gao, Qu, Zeng, Yang, Zhu, Luo, Song, Yu, Huang, Cai, Liang,
  Zhou, Ye, Li, Xu, Hu, Yang, Chen, Yan, Chen, Zhou, Xiang, Yuan, Cheng, Zhou,
  Zhu, Yu, Sun, Ran, Jiang, Qiu, Li, Zheng, Song, Dong, Gao, Guan, Zhou, Huang,
  Yu, Wang, Zhang, Wang, Xia, Zhang, Zhao, Guo, Luo, Ma, Zhu, Wang, Cai, Zhang,
  Chen, Di, Xu, Mei, Wang, Zhang, Zhang, Tang, Li, Zhou, Han, Wang, Huang,
  Wang, Cong, Wang, Zhang, Wang, Zhu, Li, Chen, Du, Jiang, Tian, Xu, Lu, Xu,
  Ge, Zhang, Pan, Wang, Chen, Yin, Xu, Shen, Zhang, Chen, Liu, Lu, Sun, Zhou,
  Chen, Cai, Nie, Wu, Chen, Hu, Liu, Hu, Ma, Wang, Yu, Zhou, Pan, Yu, Zhou, Ni,
  Yun, Jin, Pei, Ye, Lin, Ji, Cui, Yue, Yu, Wang, Zhang, Xiao, Zeng, An, Zhao,
  Liu, Liang, Pang, Luo, Yao, Gao, Yang, Huang, Hou, Zhang, Ma, Gao, He, Wang,
  Wang, Bi, Liu, Wang, Chen, Zhang, Nie, Sun, Wang, Cheng, Liu, Xie, Liu, Liu,
  Yu, Li, Yang, Zhang, Chen, Wang, Su, Chen, Lin, Fu, Yan, Wang, Ma, Luo,
  Zhang, Xu, Ma, Huang, Li, Li, Xu, Zhao, Sun, Wang, Qian, Shao, Yu, Zhang,
  Ding, Shi, Wu, Xiong, Ma, He, Tang, Zhou, Luo, Zhong, Piao, Wang, Zhang,
  Chen, Tan, Wei, Ma, Liu, Yang, Guo, Wu, Wu, Li, Cheng, Ou, Xu, Li, Wang,
  Yang, Xu, Wu, Meng, Zou, Zha, Xiong, Chen, Lin, Cao, Wang, Zhang, Yan, Lin,
  Gu, Luo, You, Liu, Zhou, Zhou, Huang, Wu, Wang, Zhao, Ren, Zhang, Sha, Fu,
  Ju, Xu, Xie, Zhang, Gao, Hao, Gou, Ma, Yan, Shao, Huang, Chen, Wu, Ren, Wu,
  Li, Zhang, Xu, Wang, Qu, Gu, Zhu, Li, Zhang, Xie, Gao, Wan, Pan, and
  Yao]{deepseekai2026v4}
DeepSeek-AI, Anyi Xu, Bangcai Lin, Bing Xue, Bingxuan Wang, Bingzheng Xu,
  Bochao Wu, Bowei Zhang, Chaofan Lin, Chen Dong, Chenchen Ling, Chengda Lu,
  Chenggang Zhao, Chengqi Deng, Chengyu Hou, Chenhao Xu, Chenze Shao, Chong
  Ruan, Conner Sun, Damai Dai, Daya Guo, Dejian Yang, Deli Chen, Donghao Li,
  Dongjie Ji, Erhang Li, Fang Wei, Fangyun Lin, Fangzhou Yuan, Feiyu Xia,
  Fucong Dai, Guangbo Hao, Guanting Chen, Guoai Cao, Guolai Meng, Guowei Li,
  Han Yu, Han Zhang, Hanwei Xu, Hao Li, Haofen Liang, Haoling Zhang, Haoming
  Luo, Haoran Wei, Haotian Yuan, Haowei Zhang, Haowen Luo, Haoyu Chen, Haozhe
  Ji, Hengqing Zhang, Honghui Ding, Hongxuan Tang, Huanqi Cao, Huazuo Gao, Hui
  Qu, Hui Zeng, J~Yang, JQ~Zhu, Jia Luo, Jia Song, Jia Yu, Jialiang Huang,
  Jialu Cai, Jian Liang, Jiangting Zhou, Jiasheng Ye, Jiashi Li, Jiaxin Xu,
  Jiewen Hu, Jieyu Yang, Jin Chen, Jin Yan, Jingchang Chen, Jingli Zhou,
  Jingting Xiang, Jingyang Yuan, Jingyuan Cheng, Jingzi Zhou, Jinhua Zhu,
  Jiping Yu, Joseph Sun, Jun Ran, Junguang Jiang, Junjie Qiu, Junlong Li,
  Junmin Zheng, Junxiao Song, Kai Dong, Kaige Gao, Kang Guan, Kexing Zhou,
  Kezhao Huang, Kuai Yu, Lean Wang, Lecong Zhang, Lei Wang, Leyi Xia, Li~Zhang,
  Liang Zhao, Lihua Guo, Lingxiao Luo, Linwang Ma, Linyan Zhu, Litong Wang,
  Liyu Cai, Liyue Zhang, Longhao Chen, MS~Di, MY~Xu, Max Mei, Miaojun Wang,
  Mingchuan Zhang, Minghua Zhang, Minghui Tang, Mingming Li, Mingxu Zhou,
  Minmin Han, Ning Wang, Panpan Huang, Panpan Wang, Peixin Cong, Peiyi Wang,
  Peng Zhang, Qiancheng Wang, Qihao Zhu, Qingyang Li, Qinyu Chen, Qiushi Du,
  Qiwei Jiang, Rui Tian, Ruifan Xu, Ruijie Lu, Ruiling Xu, Ruiqi Ge, Ruisong
  Zhang, Ruizhe Pan, Runji Wang, Runqian Chen, Runqiu Yin, Runxin Xu, Ruomeng
  Shen, Ruoyu Zhang, Ruyi Chen, SH~Liu, Shanghao Lu, Shangmian Sun, Shangyan
  Zhou, Shanhuang Chen, Shaofei Cai, Shaoheng Nie, Shaoqing Wu, Shaoyuan Chen,
  Shengding Hu, Shengyu Liu, Shiqiang Hu, Shirong Ma, Shiyu Wang, Shuiping Yu,
  Shunfeng Zhou, Shuting Pan, Shuying Yu, Songyang Zhou, Tao Ni, Tao Yun, Tian
  Jin, Tian Pei, Tian Ye, Tianle Lin, Tianran Ji, Tianyi Cui, Tianyuan Yue,
  Tingting Yu, Tun Wang, W~Zhang, WL~Xiao, Wangding Zeng, Wei An, Weilin Zhao,
  Wen Liu, Wenfeng Liang, Wenjie Pang, Wenjing Luo, Wenjing Yao, Wenjun Gao,
  Wenkai Yang, Wenlve Huang, Wenqing Hou, Wentao Zhang, Wenting Ma, Xi~Gao,
  Xiang He, Xiangwen Wang, Xianzu Wang, Xiao Bi, Xiaodong Liu, Xiaohan Wang,
  Xiaokang Chen, Xiaokang Zhang, Xiaotao Nie, Xiaowen Sun, Xiaoxiang Wang, Xin
  Cheng, Xin Liu, Xin Xie, Xingchao Liu, Xingchen Liu, Xingkai Yu, Xingyou Li,
  Xinyu Yang, Xinyu Zhang, Xu~Chen, Xuanyu Wang, Xuecheng Su, Xueyin Chen,
  Xuheng Lin, Xuwei Fu, YC~Yan, YQ~Wang, YW~Ma, Yanfeng Luo, Yang Zhang,
  Yanhong Xu, Yanru Ma, Yanwen Huang, Yao Li, Yao Li, Yao Xu, Yao Zhao, Yaofeng
  Sun, Yaohui Wang, Yi~Qian, Yi~Shao, Yi~Yu, Yichao Zhang, Yifan Ding, Yifan
  Shi, Yijia Wu, Yiliang Xiong, Yiling Ma, Ying He, Ying Tang, Ying Zhou,
  Yingjia Luo, Yinmin Zhong, Yishi Piao, Yisong Wang, Yixiang Zhang, Yixiao
  Chen, Yixuan Tan, Yixuan Wei, Yiyang Ma, Yiyuan Liu, Yonglun Yang, Yongqiang
  Guo, Yongtong Wu, Yu~Wu, YuKun Li, Yuan Cheng, Yuan Ou, Yuanfan Xu, Yuanhao
  Li, Yuduan Wang, Yuehan Yang, Yuer Xu, Yuhan Wu, Yuhao Meng, Yuheng Zou,
  Yukun Zha, Yunfan Xiong, Yupeng Chen, Yuping Lin, Yuqian Cao, Yuqian Wang,
  Yushun Zhang, Yuting Yan, Yutong Lin, Yuxian Gu, Yuxiang Luo, Yuxiang You,
  Yuxuan Liu, Yuxuan Zhou, Yuyang Zhou, Yuzhen Huang, ZF~Wu, Zehao Wang, Zehua
  Zhao, Zehui Ren, Zekai Zhang, Zhangli Sha, Zhe Fu, Zhe Ju, Zhean Xu, Zhenda
  Xie, Zhengyan Zhang, Zheren Gao, Zhewen Hao, Zhibin Gou, Zhicheng Ma, Zhigang
  Yan, Zhihong Shao, Zhixian Huang, Zhixuan Chen, Zhiyu Wu, Zhizhou Ren,
  Zhongyu Wu, Zhuoshu Li, Zhuping Zhang, Zian Xu, Zihao Wang, Zihua Qu, Zihui
  Gu, Zijia Zhu, Zilin Li, Zipeng Zhang, Ziwei Xie, Ziyi Gao, Ziyi Wan, Zizheng
  Pan, and Zongqing Yao.
\newblock Deepseek-v4: Towards highly efficient million-token context
  intelligence, 2026.
\newblock URL \url{https://arxiv.org/abs/2606.19348}.

\bibitem[Dou et~al.(2026{\natexlab{a}})Dou, Shen, Huang, Ye, Chen, Wang, He,
  Liu, Lv, Lin, Zhang, Zhang, Liu, Ji, Yin, Zhang, Xie, Hu, Deng, Li, Hu, Wang,
  Zhao, Wang, Lei, Liu, Xiao, Liu, Xu, Guo, Zhao, Zhou, Gui, Zhang, Huang,
  Jiang, Wang, and Yao]{dou2026clbenchlife}
Shihan Dou, Yujiong Shen, Chenhao Huang, Junjie Ye, Jiayi Chen, Junzhe Wang,
  Qianyu He, Shichun Liu, Changze Lv, Jiahang Lin, Jiazheng Zhang, Ming Zhang,
  Shaofan Liu, Tao Ji, Zhangyue Yin, Cheng Zhang, Huaibing Xie, Jianglu Hu,
  Jingcheng Deng, Lincheng Li, Minda Hu, Shaolei Wang, Syrus Zhao, Weichao
  Wang, Yan Lei, Yang Liu, Yanling Xiao, Yiting Liu, Zenan Xu, Zhen Guo,
  Ziliang Zhao, Pluto Zhou, Tao Gui, Qi~Zhang, Xuanjing Huang, Yu-Gang Jiang,
  Di~Wang, and Shunyu Yao.
\newblock {CL-bench Life}: Can language models learn from real-life context?,
  2026{\natexlab{a}}.
\newblock URL \url{https://arxiv.org/abs/2604.27043}.

\bibitem[Dou et~al.(2026{\natexlab{b}})Dou, Zhang, Yin, Huang, Shen, Wang,
  Chen, Ni, Ye, Zhang, Xie, Hu, Wang, Wang, Xiao, Liu, Xu, Guo, Zhou, Gui, Wu,
  Qiu, Zhang, Huang, Jiang, Wang, and Yao]{dou2026clbench}
Shihan Dou, Ming Zhang, Zhangyue Yin, Chenhao Huang, Yujiong Shen, Junzhe Wang,
  Jiayi Chen, Yuchen Ni, Junjie Ye, Cheng Zhang, Huaibing Xie, Jianglu Hu,
  Shaolei Wang, Weichao Wang, Yanling Xiao, Yiting Liu, Zenan Xu, Zhen Guo,
  Pluto Zhou, Tao Gui, Zuxuan Wu, Xipeng Qiu, Qi~Zhang, Xuanjing Huang, Yu-Gang
  Jiang, Di~Wang, and Shunyu Yao.
\newblock {CL-bench}: A benchmark for context learning, 2026{\natexlab{b}}.
\newblock URL \url{https://arxiv.org/abs/2602.03587}.

\bibitem[Geng et~al.(2022)Geng, Liu, Fu, Ge, and Zhang]{geng2022p5}
Shijie Geng, Shuchang Liu, Zuohui Fu, Yingqiang Ge, and Yongfeng Zhang.
\newblock Recommendation as language processing ({RLP}): A unified pretrain,
  personalized prompt \& predict paradigm ({P5}).
\newblock In \emph{Proceedings of the 16th ACM Conference on Recommender
  Systems}, RecSys '22, pp.\  299--315. ACM, 2022.
\newblock \doi{10.1145/3523227.3546767}.
\newblock URL \url{http://dx.doi.org/10.1145/3523227.3546767}.

\bibitem[Hong et~al.(2026)Hong, Zeng, Lei, Wang, Zhang, Jiang, Fu, Huang, Liu,
  Chang, Zou, Zhou, He, Simonjmdeng, Huang, Li, Li, Zhang, Zhou, Ma, Hua, Zhu,
  Jiang, Ma, Zhang, Song, Peng, Liang, and Chen]{uxbench2026}
Mengze Hong, Xia Zeng, Zeyang Lei, Sheng Wang, Chen~Jason Zhang, Di~Jiang,
  Taiming Fu, Jinfeng Huang, Mengqiao Liu, Qinghe Chang, Haosheng Zou, Qiongyi
  Zhou, Sijun He, Simonjmdeng, Haojing Huang, Zijian Li, Lucas~Mu Li, Fubao
  Zhang, Mona Zhou, Wei Ma, Yuan Hua, Qi~Zhu, Shuo Jiang, Chenxuan Ma, Yuanmeng
  Zhang, Jian Song, Minlong Peng, Di~Liang, and Davey Chen.
\newblock {UXBench}: Benchmarking user experience in {AI} assistants, 2026.
\newblock URL \url{https://arxiv.org/abs/2606.09570}.

\bibitem[Jiang et~al.(2025)Jiang, Hao, Cho, Li, Yuan, Chen, Ungar, Taylor, and
  Roth]{song2025personamem}
Bowen Jiang, Zhuoqun Hao, Young-Min Cho, Bryan Li, Yuan Yuan, Sihao Chen, Lyle
  Ungar, Camillo~J. Taylor, and Dan Roth.
\newblock Know me, respond to me: Benchmarking {LLMs} for dynamic user
  profiling and personalized responses at scale, 2025.
\newblock URL \url{https://arxiv.org/abs/2504.14225}.

\bibitem[Li et~al.(2026)Li, Wang, Gu, Wang, Zhuang, and Bian]{liveagent2026}
Hao Li, Huan Wang, Jinjie Gu, Wenjie Wang, Chenyi Zhuang, and Sikang Bian.
\newblock {LiveAgentBench}: Comprehensive benchmarking of agentic systems
  across 104 real-world challenges, 2026.
\newblock URL \url{https://arxiv.org/abs/2603.02586}.

\bibitem[Li et~al.(2025)Li, Dai, Zhou, and Hu]{li2025finerob}
Kun Li, Chengwei Dai, Wei Zhou, and Songlin Hu.
\newblock {LLM}-based behavior prediction for social media users with
  continuous memory.
\newblock In Kentaro Inui, Sakriani Sakti, Haofen Wang, Derek~F. Wong, Pushpak
  Bhattacharyya, Biplab Banerjee, Asif Ekbal, Tanmoy Chakraborty, and
  Dhirendra~Pratap Singh (eds.), \emph{Proceedings of the 14th International
  Joint Conference on Natural Language Processing and the 4th Conference of the
  Asia-Pacific Chapter of the Association for Computational Linguistics}, pp.\
  459--474, Mumbai, India, 2025. The Asian Federation of Natural Language
  Processing and The Association for Computational Linguistics.
\newblock ISBN 979-8-89176-298-5.
\newblock \doi{10.18653/v1/2025.ijcnlp-long.27}.
\newblock URL \url{https://aclanthology.org/2025.ijcnlp-long.27/}.

\bibitem[Lozhkov et~al.(2026)Lozhkov, Larcher, Morlon, Ben~Allal, and von
  Werra]{stackv32026}
Anton Lozhkov, Hugo Larcher, Mathieu Morlon, Loubna Ben~Allal, and Leandro von
  Werra.
\newblock The {Stack v3}: The largest open code dataset, 2026.
\newblock URL
  \url{https://huggingface.co/datasets/HuggingFaceCode/stack-v3-train}.

\bibitem[{MAA}(2026)]{aime2026}
{MAA}.
\newblock American invitational mathematics examination ({AIME}) 2026: {I} and
  {II}.
\newblock Mathematical Association of America, 2026.
\newblock URL \url{https://live.poshenloh.com/past-contests/aime/2026I}.
\newblock MAA-authorized reproduction. AIME II:
  \url{https://live.poshenloh.com/past-contests/aime/2026II}. Accessed
  2026-09-26.

\bibitem[Mei et~al.(2026)Mei, Chen, Yang, Hou, Li, and Byrne]{mei2026atmbench}
Jingbiao Mei, Jinghong Chen, Guangyu Yang, Xinyu Hou, Margaret Li, and Bill
  Byrne.
\newblock According to me: Long-term personalized referential memory {QA},
  2026.
\newblock URL \url{https://arxiv.org/abs/2603.01990}.

\bibitem[Ning et~al.(2024)Ning, Liu, Wu, Wu, Berlowitz, Prakash, Green,
  O'Banion, and Xie]{userllm2025}
Lin Ning, Luyang Liu, Jiaxing Wu, Neo Wu, Devora Berlowitz, Sushant Prakash,
  Bradley Green, Shawn O'Banion, and Jun Xie.
\newblock {User-LLM}: Efficient {LLM} contextualization with user embeddings,
  2024.
\newblock URL \url{https://arxiv.org/abs/2402.13598}.

\bibitem[Pyatkin et~al.(2025)Pyatkin, Malik, Graf, Ivison, Huang, Dasigi,
  Lambert, and Hajishirzi]{ifbench2025}
Valentina Pyatkin, Saumya Malik, Victoria Graf, Hamish Ivison, Shengyi Huang,
  Pradeep Dasigi, Nathan Lambert, and Hannaneh Hajishirzi.
\newblock Generalizing verifiable instruction following, 2025.
\newblock URL \url{https://arxiv.org/abs/2507.02833}.

\bibitem[{Qwen Team}(2026{\natexlab{a}})]{qwen35397ba17b}
{Qwen Team}.
\newblock {Qwen3.5-397B-A17B}, 2026{\natexlab{a}}.
\newblock URL \url{https://huggingface.co/Qwen/Qwen3.5-397B-A17B}.

\bibitem[{Qwen Team}(2026{\natexlab{b}})]{qwen3635ba3b}
{Qwen Team}.
\newblock {Qwen3.6-35B-A3B}: Agentic coding power, now open to all, April
  2026{\natexlab{b}}.
\newblock URL \url{https://qwen.ai/blog?id=qwen3.6-35b-a3b}.
\newblock Model card: \url{https://huggingface.co/Qwen/Qwen3.6-35B-A3B}.

\bibitem[{Qwen Team}(2026{\natexlab{c}})]{qwenlm2026qwen38}
{Qwen Team}.
\newblock {Qwen3.8-Max}: A new bar for coding and cowork, August
  2026{\natexlab{c}}.
\newblock URL \url{https://qwen.ai/blog?id=qwen3.8}.
\newblock Model card: \url{https://huggingface.co/Qwen/Qwen3.8-2.4T-A95B}.
  Accessed 2026-09-26.

\bibitem[Rein et~al.(2023)Rein, Hou, Stickland, Petty, Pang, Dirani, Michael,
  and Bowman]{gpqa2023}
David Rein, Betty~Li Hou, Asa~Cooper Stickland, Jackson Petty, Richard~Yuanzhe
  Pang, Julien Dirani, Julian Michael, and Samuel~R. Bowman.
\newblock {GPQA}: A graduate-level google-proof {Q\&A} benchmark, 2023.
\newblock URL \url{https://arxiv.org/abs/2311.12022}.

\bibitem[Shumailov et~al.(2024)Shumailov, Shumaylov, Zhao, Papernot, Anderson,
  and Gal]{shumailov2024ai}
Ilia Shumailov, Zakhar Shumaylov, Yiren Zhao, Nicolas Papernot, Ross Anderson,
  and Yarin Gal.
\newblock {AI} models collapse when trained on recursively generated data.
\newblock \emph{Nature}, 631\penalty0 (8022):\penalty0 755--759, 2024.
\newblock ISSN 1476-4687.
\newblock \doi{10.1038/s41586-024-07566-y}.
\newblock URL \url{http://dx.doi.org/10.1038/s41586-024-07566-y}.

\bibitem[Steinberger \& {OpenClaw Contributors}(2025)Steinberger and {OpenClaw
  Contributors}]{openclaw2025}
P.~Steinberger and {OpenClaw Contributors}.
\newblock {OpenClaw}, 2025.
\newblock \url{https://github.com/openclaw/openclaw}.

\bibitem[Su et~al.(2025)Su, Kong, Lin, Jennings, Norick, Kliegl, Patwary,
  Shoeybi, and Catanzaro]{su2025nemotroncc}
Dan Su, Kezhi Kong, Ying Lin, Joseph Jennings, Brandon Norick, Markus Kliegl,
  Mostofa Patwary, Mohammad Shoeybi, and Bryan Catanzaro.
\newblock Nemotron-{CC}: Transforming {C}ommon {C}rawl into a refined
  long-horizon pretraining dataset.
\newblock In Wanxiang Che, Joyce Nabende, Ekaterina Shutova, and Mohammad~Taher
  Pilehvar (eds.), \emph{Proceedings of the 63rd Annual Meeting of the
  Association for Computational Linguistics (Volume 1: Long Papers)}, pp.\
  2459--2475, Vienna, Austria, July 2025. Association for Computational
  Linguistics.
\newblock ISBN 979-8-89176-251-0.
\newblock \doi{10.18653/v1/2025.acl-long.123}.
\newblock URL \url{https://aclanthology.org/2025.acl-long.123/}.

\bibitem[Team et~al.(2026{\natexlab{a}})Team, Abd, Aggarwal, Algayres, Andreev,
  Bachem, Ballantyne, Brick, C\u{a}rbune, Casbon, Chaturvedi, Chawla, Cotruta,
  Coucke, Culliton, Dadashi, Dixon, Elhawaty, Evci, Farabet, Ferret, Galgani,
  Girgin, Grill, Grootendorst, Guo, Hardin, He, Hernandez, Homburger, Hussenot,
  Ji, Joulin, Kamath, Kassraie, Lacombe, Lahoti, Liu, Martins, Martins,
  Matejovicova, Merhej, Momchev, Mondal, Mullins, Panyam, Pathak, Perrin,
  Pinto, Pot, Pouget, Ramé, Ramos, Reid, Rim, Rivière, Roth, Rouillard,
  Sanseviero, Sessa, Settle, Sinopalnikov, Smoot, Stanczyk, Steiner, Stewart,
  Tolstikhin, Tschannen, Tsitsulin, Vieillard, Wu, Xu, Yang, Yvinec, Zhang,
  Zhang, Zou, Aagnes, Abdelhamed, Adamek, Agrawal, Agrawal, Alabdulmohsin,
  Alayrac, Alon, Amarnath, Anand, Anastasiou, Ariafar, Aubet, Axiotis, Barbero,
  Barral, Bendebury, Bergmann, Bileschi, Black, Blondel, Borgeaud,
  Bra\v{z}inskas, Burnell, Busa-Fekete, Cai, Calandriello, Cameron, Caucheteux,
  Chaabouni, Chadha, Chan, Chen, Chen, Chen, Chen, Cheng, hsiang Chien,
  Chinaev, Chou, Chu, Coleman, Consul, Conway-Rahman, Crowell, Cutler, Dani,
  Daruki, Das, Deutsch, Dikkala, Ding, Ding, Dodhia, Donhauser, Doshi, Dragan,
  Druinsky, Dua, Egyed, Eisenbud, Eppens, Fan, Fatemi, Fathullah, Feinberg,
  Ferev, Flennerhag, Fujimoto, Oliveira, Galatzer-Levy, Gante, Geisler, Ghosal,
  Girgis, von Glehn, Go, Gokhale, Grills, Gu, Gupta, Gupta, Guruganesh,
  Hadsell, Harkous, Harlalka, Hassabis, Hauth, Heyward, Hosseini, Hsia, Hsu,
  Huang, Huang, Hui, Hutter, I, Iliopoulos, Jain, Jawahar, Ji, Jin, Johnson,
  Joshi, Kandoor, Kang, Kavukcuoglu, Kazemi, Kenealy, Khalifa, Kirk, Korotkov,
  Kothawade, Kovalev, Kovelamudi, Kraft, Kumar, Kumar, Kuppam, Lannin, Lee,
  Lee, Lepikhin, Levkovitch, Li, Li, Liévin, Lin, Lin, Liu, Liu, Liu, Liu,
  Lobov, Lunayach, Ma, Madan, Maksai, Malmi, Matuszak, McDuff, Menghani,
  Miku\l{}a, Mirylenka, Misiunas, Misra, Mitran, Mohamed, Mukha, Noland,
  O'Donnell, O'Donoghue, Olszewska, Orlando, Pan, Panigrahy, Parekh,
  Perez-Nieves, Park, Paskie, Peng, Petrini, Petrov, Pfeiffer, Piot, Plomecka,
  Poder, Ponce, Pramanik, Racz, Rajan, Ramanovich, Rao, Ritter, Rodrigues,
  Rosen, Rybi\'{n}ski, Sachdeva, Sander, Sathyanarayana, Savla, Schmidgall,
  Schuster, Scrivener, Seguin, Sellergren, Severyn, Shafran, Shah, Shahriari,
  Shangguan, Shenoy, Shenoy, Shivanna, Sho, Spangher, Stokowiec, Strother, Su,
  Sun, Sundararajan, Tacchetti, Taege, Tafti, Tarbouriech, Tekur, Thakoor,
  Thapa, Traverse, Treven, Tu, Tung, Ça\u{g}lar Ünlü, Veli\v{c}kovi\'{c},
  Venkat, Venkatesh, Venkiteswaran, Visin, Vitvitskyi, Vodrahalli, Wang, Wang,
  Warkentin, Wassenberg, Wieting, Wu, Xiao, Xu, Xu, Xue, Yadav, Yan, Yang,
  Yang, Yang, Ying, Yoo, Zadimoghaddam, Zafar, Zhang, Zhang, Zhang, Zhang,
  Zhao, Zhou, and Zou]{gemmateam2026gemma4}
Gemma Team, Sherif~El Abd, Vaibhav Aggarwal, Robin Algayres, Alek Andreev,
  Olivier Bachem, Ian Ballantyne, Cormac Brick, Victor C\u{a}rbune, Michelle
  Casbon, Mayank Chaturvedi, Aditya Chawla, Victor Cotruta, Alice Coucke, Phil
  Culliton, Robert Dadashi, Lucas Dixon, Mohamed Elhawaty, Utku Evci, Clément
  Farabet, Johan Ferret, Filippo Galgani, Sertan Girgin, Jean-Bastien Grill,
  Maarten Grootendorst, Jiaxian Guo, Cassidy Hardin, Yanzhang He, Steven~M.
  Hernandez, Omri Homburger, Léonard Hussenot, Juyeong Ji, Armand Joulin,
  Aishwarya Kamath, Parnian Kassraie, Olivier Lacombe, Preethi Lahoti, Gaël
  Liu, Gus Martins, Luciano Martins, Tatiana Matejovicova, Ramona Merhej,
  Nikola Momchev, Sneha Mondal, Ryan Mullins, Sindhu~Raghuram Panyam, Shreya
  Pathak, Sarah Perrin, André~Susano Pinto, Etienne Pot, Angéline Pouget,
  Alexandre Ramé, Sabela Ramos, Douglas Reid, David Rim, Morgane Rivière,
  Karsten Roth, Louis Rouillard, Omar Sanseviero, Pier~Giuseppe Sessa, Shane
  Settle, Danila Sinopalnikov, Sara Smoot, Piotr Stanczyk, Andreas Steiner,
  Lawrence Stewart, Ilya Tolstikhin, Michael Tschannen, Anton Tsitsulin, Nino
  Vieillard, Renjie Wu, Pingmei Xu, Haichuan Yang, Edouard Yvinec, Biao Zhang,
  Li~Zhang, Joe Zou, Nicolas Aagnes, Abdelrahman Abdelhamed, Jakub Adamek,
  Shivani Agrawal, Shubham Agrawal, Ibrahim Alabdulmohsin, Jean~Baptiste
  Alayrac, Uri Alon, Chandramouli Amarnath, Ankesh Anand, Chrysovalantis
  Anastasiou, Setareh Ariafar, François-Xavier Aubet, Kyriakos Axiotis,
  Federico Barbero, Joelle Barral, Alexei Bendebury, Urs Bergmann, Stanley
  Bileschi, Kat Black, Mathieu Blondel, Sebastian Borgeaud, Arthur
  Bra\v{z}inskas, Ryan Burnell, Robert Busa-Fekete, Mu~Cai, Daniele
  Calandriello, Glenn Cameron, Charlotte Caucheteux, Rahma Chaabouni, Garima
  Chadha, Jetha Chan, Blake~Jianhang Chen, Jesse Chen, Lin Chen, Xu~Chen, Derek
  Cheng, Tzu hsiang Chien, Nikolai Chinaev, Yi~Chou, Zhaohui Chu, Benjamin
  Coleman, Pooja Consul, Sam Conway-Rahman, Scott Crowell, Dylan Cutler, Vivek
  Dani, Samira Daruki, Anil Das, Daniel Deutsch, Nishanth Dikkala, Li~Ding,
  Qiuhan Ding, Shenil Dodhia, Konstantin Donhauser, Tulsee Doshi, Anca Dragan,
  Alex Druinsky, Sahil Dua, Zoltan Egyed, Danielle Eisenbud, Daniel Eppens,
  Cindy Fan, Bahare Fatemi, Yassir Fathullah, Vlad Feinberg, Milen Ferev,
  Sebastian Flennerhag, Takumi Fujimoto, João~Gabriel Oliveira, Isaac
  Galatzer-Levy, João Gante, Simon Geisler, Soham Ghosal, Antonious~M. Girgis,
  Tamara von Glehn, Alec Go, Alhaad Gokhale, Alex Grills, Yiming Gu, Mayank
  Gupta, Pramod Gupta, Guru Guruganesh, Raia Hadsell, Hamza Harkous, Jitendra
  Harlalka, Demis Hassabis, Anja Hauth, Joe Heyward, Arian Hosseini, Chih-Yang
  Hsia, I-Hung Hsu, Xiaopeng Huang, Yangsibo Huang, Kevin Hui, Adrian Hutter,
  Te~I, Fotis Iliopoulos, Advait Jain, Ganesh Jawahar, Ziwei Ji, Qilin Jin,
  Melvin Johnson, Kandarp Joshi, Arun Kandoor, Wang-Cheng Kang, Koray
  Kavukcuoglu, Mehran Kazemi, Kathleen Kenealy, Amr Khalifa, Phoebe Kirk, Ivan
  Korotkov, Suraj Kothawade, Vitaly Kovalev, Neel Kovelamudi, Adam Kraft, Ravin
  Kumar, Vivek Kumar, Harish Kuppam, Justin Lannin, Chen-Yu Lee, Seungji Lee,
  Dmitry Lepikhin, Alon Levkovitch, Dongdong Li, Qiujia Li, Valentin Liévin,
  Ethan Lin, Ziqian Lin, Casper Liu, Tianlin Liu, Tianqi Liu, Xin Liu, Ivan
  Lobov, Mayank Lunayach, Min Ma, Gagan Madan, Andrii Maksai, Eric Malmi,
  Michal Matuszak, Daniel McDuff, Gaurav Menghani, Maciej Miku\l{}a, Daniil
  Mirylenka, Karolis Misiunas, Vedant Misra, Andreea Mitran, Kareem Mohamed,
  Maksim Mukha, Eric Noland, James O'Donnell, Brendan O'Donoghue, Kate
  Olszewska, Bernett Orlando, Wanqiong Pan, Rina Panigrahy, Unnati Parekh,
  Nicolas Perez-Nieves, Chunjong Park, Eric Paskie, Liqian Peng, Bryce Petrini,
  Slav Petrov, Jonas Pfeiffer, Bilal Piot, Martyna Plomecka, Siim Poder,
  Octavio Ponce, Arijit Pramanik, David Racz, Anish Rajan, Michelle Ramanovich,
  Anand Rao, Marvin Ritter, Vitor Rodrigues, Evan Rosen, Miko\l{}aj
  Rybi\'{n}ski, Noveen Sachdeva, Michaël~E. Sander, Rohit Sathyanarayana,
  Sagar Savla, Samuel Schmidgall, Tal Schuster, George Scrivener, Benoit
  Seguin, Andrew Sellergren, Aliaksei Severyn, Izhak Shafran, Dhruv Shah, Bobak
  Shahriari, Yuan Shangguan, Ashish Shenoy, Pradeep Shenoy, Rakesh Shivanna,
  Pauline Sho, Lucas Spangher, Wojciech Stokowiec, Tim Strother, Yao Su,
  Yinghao Sun, Mukund Sundararajan, Andrea Tacchetti, Mor~Hazan Taege, Pouya
  Tafti, Jean Tarbouriech, Chetan Tekur, Shantanu Thakoor, Rahul Thapa,
  Madeleine Traverse, Lenart Treven, Tao Tu, Chien~Te Tung, Ça\u{g}lar Ünlü,
  Petar Veli\v{c}kovi\'{c}, Malini~Pooni Venkat, Sagar~Gubbi Venkatesh, Vidya
  Venkiteswaran, Francesco Visin, Alex Vitvitskyi, Kiran Vodrahalli, Weiyi
  Wang, Xin Wang, Tris Warkentin, Jan Wassenberg, John Wieting, Cindy Wu,
  Lechao Xiao, Hao Xu, Yuhui Xu, Fuzhao Xue, Arun Yadav, Jun Yan, Antoine Yang,
  Lin Yang, Ming-Hsuan Yang, Ziyu Ying, Jae~Hyeon Yoo, Morteza Zadimoghaddam,
  Sajjad Zafar, Fred Zhang, Jiageng Zhang, Jianyi Zhang, Xiaofan Zhang, Chao
  Zhao, David Zhou, and Chen Zou.
\newblock Gemma 4 technical report, 2026{\natexlab{a}}.
\newblock URL \url{https://arxiv.org/abs/2607.02770}.

\bibitem[Team et~al.(2026{\natexlab{b}})Team, Bai, Bai, Bao, C., Cai, Cai, Cao,
  Cao, Chai, Charles, Che, Chen, Chen, Chen, Chen, Chen, Chen, Chen, Chen,
  Chen, Chen, Chen, Chen, Chen, Chen, Chen, Chen, Chen, Chen, Chen, Chen,
  Cheng, Cheng, Cui, Cui, Dai, Deng, Ding, Ding, Ding, Dong, Dong, Dong, Dong,
  Du, Du, Du, Du, Du, Fan, Feng, Feng, Feng, Fu, Fu, Gao, Gao, Gao, Gao, Gao,
  Geng, Gong, Gong, Gong, Gong, Gu, Gu, Guan, Guo, Guo, Guo, Guo, Hao, Hao,
  Hao, He, He, He, He, He, He, He, He, He, Hong, Hong, Hu, Hu, Hu, Hu, Hu, Hu,
  Hua, Huang, Huang, Huang, Huang, Huang, Huang, Huang, Huang, Hui, Jia, Jiang,
  Jiang, Jiang, Jin, Jin, Jing, Kong, Lai, Li, Li, Li, Li, Li, Li, Li, Li, Li,
  Li, Li, Li, Li, Li, Li, Li, Li, Li, Li, Li, Li, Li, Li, Li, Lin, Lin, Lin,
  Lin, Lin, Liu, Liu, Liu, Liu, Liu, Liu, Liu, Liu, Liu, Liu, Liu, Liu, Liu,
  Liu, Liu, Lu, Lu, Lu, Lu, Lu, Luo, Luo, Luo, Luo, Lyu, Lyu, Mao, Mei, Men,
  Ni, Niu, Pan, Peng, Qi, Qin, Qin, Qin, Qiu, Qiu, Qiu, Qu, Qu, Shang, Shao,
  Shen, Shi, Shi, Shi, Shi, Siu, Song, Song, Su, Su, Su, Sui, Sun, Sun, Sun,
  Sun, Sun, Sun, Tai, Tang, Tang, Tang, Tang, Tian, Tian, Tian, Tu, Wang, Wang,
  Wang, Wang, Wang, Wang, Wang, Wang, Wang, Wang, Wang, Wang, Wang, Wang, Wang,
  Wang, Wang, Wang, Wang, Wang, Wang, Wang, Wang, Wang, Wang, Wang, Wang, Wang,
  Wang, Wang, Wang, Wang, Wang, Wang, Wang, Wang, Wang, Wang, Wang, Wang, Wang,
  Wei, Wei, Wei, Wen, Wu, Wu, Wu, Wu, Wu, Wu, Wu, Wu, Wu, Xian, Xiang, Xiang,
  Xiao, Xiao, Xiao, Xie, Xie, Xie, Xie, Xing, Xiong, Xu, Xu, Xu, Xu, Xu, Xu,
  Xu, Xu, Xu, Xu, Xu, Xu, Xu, Xu, Xu, Xu, Xu, Xu, Xue, Yan, Yan, Yang, Yang,
  Yang, Yang, Yang, Yang, Yang, Yang, Yang, Yang, Yang, Yang, Yang, Yang, Yang,
  Yang, Yao, Ye, Ye, Ye, Ye, Yin, Yin, Yin, Yu, Yu, Yu, Yu, Yu, Yu, Yuan, Yuan,
  Yue, Yue, Yue, Zha, Zhan, Zhang, Zhang, Zhang, Zhang, Zhang, Zhang, Zhang,
  Zhang, Zhang, Zhang, Zhang, Zhang, Zhang, Zhang, Zhang, Zhang, Zhang, Zhang,
  Zhang, Zhang, Zhang, Zhang, Zhang, Zhang, Zhang, Zhang, Zhang, Zhang, Zhang,
  Zhang, Zhang, Zhao, Zhao, Zhao, Zhao, Zhao, Zhao, Zhao, Zhao, Zhao, Zhao,
  Zhao, Zheng, Zheng, Zheng, Zheng, Zheng, Zhong, Zhong, Zhong, Zhou, Zhou,
  Zhou, Zhou, Zhou, Zhou, Zhou, Zhu, Zhu, Zhu, Zhu, Zhu, Zhu, Zhuang, Zhuang,
  and Zu]{moonshotai2026kimik3}
Kimi Team, Tongtong Bai, Yifan Bai, Yiping Bao, M.~C., Jianfeng Cai, Xinyuan
  Cai, Peizhou Cao, Yuxuan Cao, Ziwei Chai, Y.~Charles, H.~S. Che, Guanduo
  Chen, Guangyu Chen, Guanzheng Chen, Huarong Chen, Jia Chen, Jianlong Chen,
  Jun Chen, Kexin Chen, Peng Chen, Ruijue Chen, Wentao Chen, Xin Chen, Yang
  Chen, Yanru Chen, Yifei Chen, Yingjiang Chen, Yuankun Chen, Yujie Chen,
  Yutian Chen, Zhirong Chen, Dazhi Cheng, Yean Cheng, Jialei Cui, Jingbing Cui,
  Anqi Dai, Jiaqi Deng, Hao Ding, Rui Ding, Shaofeng Ding, Mengfan Dong,
  Mengnan Dong, Yuhao Dong, Yuxin Dong, Angang Du, Chenzhuang Du, Dikang Du,
  Jusen Du, Yulun Du, Yu~Fan, Jing Feng, Qiulin Feng, Yichen Feng, Kelin Fu,
  Qiang Fu, Fuxuan Gao, Hongcheng Gao, Jingyue Gao, Tong Gao, Weijia Gao,
  Shangyi Geng, Jie Gong, Linhu Gong, Shengao Gong, Xiaochen Gong, Qizheng Gu,
  Yicheng Gu, Shuhao Guan, Haiqing Guo, Shiqi Guo, Xiang Guo, Zhengyan Guo,
  Beixi Hao, Wenxin Hao, Xiaoru Hao, Dailan He, Haotian He, Lehan He, Qi~He,
  Weiran He, Xinran He, Xinyi He, Yibo He, Yunjia He, Chao Hong, Tiange Hong,
  Hao Hu, Jiaxi Hu, Ruikun Hu, Weiming Hu, Yangyang Hu, Zhenxing Hu, Liang Hua,
  Jinbin Huang, Ke~Huang, Ruiyuan Huang, Siying Huang, Weixiao Huang, Yan
  Huang, Zhengjie Huang, Zhiqi Huang, Yulong Hui, Chaobo Jia, Yutong Jiang,
  Zhejun Jiang, Zuoyou Jiang, Wenyi Jin, Xinyi Jin, Yu~Jing, Huanjun Kong,
  Guokun Lai, Aidi Li, Cheng Li, Chengyuan Li, Cong Li, Fang Li, Guanyu Li,
  Haoyang Li, Jia Li, Junxiong Li, Lei Li, Letian Li, Lincan Li, Weihong Li,
  Wentao Li, Xintong Li, Yang Li, Yishen Li, Yiwei Li, Yuxiao Li, Zhaowei Li,
  Zhaoxi Li, Zheming Li, Zhengxiao Li, Zhiyuan Li, Jiawei Lin, Xiaohan Lin,
  Yibo Lin, Zichao Lin, Ziyan Lin, Bill Liu, Boxiao Liu, Chuan Liu, Liang Liu,
  Shaowei Liu, Shudong Liu, Shuran Liu, Tianwei Liu, Weizhou Liu, Yangyang Liu,
  Yanming Liu, Yibo Liu, Yipeng Liu, Zhengying Liu, Zhiheng Liu, Enzhe Lu,
  Haoyu Lu, Linqiang Lu, Tingzhan Lu, Zhiyuan Lu, Aotian Luo, G.~Luo, Junyu
  Luo, Yifan Luo, B.~Lyu, Wenzhou Lyu, Shaoguang Mao, Yuan Mei, Xin Men,
  Minqing Ni, Yixuan Niu, Siyuan Pan, Shujun Peng, Zhangyang Qi, Ruoyu Qin,
  ZeChao Qin, Zeyu Qin, Haiquan Qiu, Jianxin Qiu, Jiezhong Qiu, Bowen Qu, Yuhao
  Qu, Zeyu Shang, Youbo Shao, Han Shen, Jincheng Shi, Juanfeng Shi, Lidong Shi,
  Shengyuan Shi, Wingchun Siu, Pengwei Song, Xiaoxi Song, Jianlin Su, Yunfeng
  Su, Zhaochen Su, Lin Sui, Jingsong Sun, Junyao Sun, Shaoning Sun, Shuzhe Sun,
  Tongyu Sun, Yujun Sun, Yunpeng Tai, Chuning Tang, Heyi Tang, Sirui Tang,
  Zecheng Tang, Chaoran Tian, Rongpeng Tian, Yu~Tian, Wei Tu, Chensi Wang,
  Chuang Wang, Chunjie Wang, Dinglu Wang, Feng Wang, Hailong Wang, Haiming
  Wang, Hao Wang, Hao Wang, Huaqing Wang, Hui Wang, Jiayi Wang, Jinglong Wang,
  Jinhong Wang, Jiuzheng Wang, Linian Wang, Shaobo Wang, Shenzhi Wang, Shuyi
  Wang, Si~Wang, Siyuan Wang, Tianfu Wang, Wenjue Wang, Xingran Wang, Xinmei
  Wang, Xinyuan Wang, Xusheng Wang, Yalin Wang, Yangkun Wang, Yao Wang, Yaoyu
  Wang, Yejie Wang, Yiqin Wang, Yucheng Wang, Yuzhi Wang, Zhaoji Wang, Zhaowei
  Wang, Zhengtao Wang, Zhenhao Wang, Zhongsheng Wang, Zifan Wang, Chu Wei, Ming
  Wei, Shouxin Wei, Zichen Wen, Fan Wu, Haoning Wu, Rucong Wu, Wenhao Wu,
  Xiaoxue Wu, Yingcong Wu, Yongqi Wu, Yuxin Wu, Zijian Wu, Xinglang Xian,
  Chenxuan Xiang, Yuye Xiang, Bocheng Xiao, Chenjun Xiao, Xin Xiao, Jin Xie,
  Xiaotong Xie, Yifeng Xie, Zhe Xie, Bowei Xing, Yiming Xiong, Baosheng Xu,
  Boyu Xu, Jiale Xu, Jianfan Xu, Jing Xu, Jinjing Xu, L.~H. Xu, Qingtao Xu,
  Shuyao Xu, Suting Xu, Tiantian Xu, Tianxiang Xu, Weixin Xu, Xinran Xu,
  Yangchuan Xu, Ye~Xu, Yueni Xu, Ziyao Xu, Haonan Xue, Junjie Yan, Yaoyao Yan,
  Fan Yang, Guangyao Yang, Hao Yang, Junwei Yang, Ruoyu Yang, Wenjie Yang,
  Xiaofei Yang, Xinyu Yang, Yi~Yang, Yiling Yang, Ying Yang, Yuchen Yang, Zhen
  Yang, Zhilin Yang, Zian Yang, Zuhao Yang, Haotian Yao, Dan Ye, Haoran Ye,
  Wenjie Ye, Zhanbo Ye, Bohong Yin, Haoxiang Yin, Xietong Yin, Chengzhen Yu,
  Haozhen Yu, Longhui Yu, Shengnan Yu, Shuying Yu, Tianxiang Yu, Enming Yuan,
  Mengjie Yuan, Tongtian Yue, Wei Yue, Yang Yue, Dunyuan Zha, Haobing Zhan,
  B.~H. Zhang, Dehao Zhang, Fei Zhang, Hao Zhang, Haoyuan Zhang, Huanyu Zhang,
  Jiapei Zhang, Jiaxuan Zhang, Jin Zhang, Kaiyi Zhang, Miaozhen Zhang, Puqi
  Zhang, Qinglei Zhang, Rong Zhang, Rui Zhang, Shaoshuai Zhang, Shiyi Zhang,
  Xiaobin Zhang, Xiaoyun Zhang, Y.~Zhang, Yangkun Zhang, Ye~Zhang, Yichi Zhang,
  Yikun Zhang, Yizhi Zhang, Yongting Zhang, Yu~Zhang, Yutao Zhang, Yutong
  Zhang, Zheng Zhang, Zijing Zhang, Bin Zhao, Chenguang Zhao, Feifan Zhao,
  Jinglun Zhao, Jinxiang Zhao, Shuai Zhao, Wenshuo Zhao, Xiangyu Zhao, Xuanle
  Zhao, Yikai Zhao, Zijia Zhao, Haozhi Zheng, Huabin Zheng, Ruihan Zheng,
  Shaojie Zheng, Tengyang Zheng, Haofeng Zhong, Lei Zhong, Longguang Zhong,
  M.~Zhou, Qiankang Zhou, Runjie Zhou, Ruozhang Zhou, Xinyu Zhou, Yiqiao Zhou,
  Zaida Zhou, Jinguo Zhu, Liya Zhu, Xinhao Zhu, Yangjunfeng Zhu, Yuxuan Zhu,
  Zhen Zhu, Chen Zhuang, Weiyu Zhuang, and Xinxing Zu.
\newblock Kimi k3: Open frontier intelligence, 2026{\natexlab{b}}.
\newblock URL \url{https://arxiv.org/abs/2607.24653}.

\bibitem[Uddin et~al.(2026)Uddin, Shubham, Blanco, Baral, and Wang]{memora2026}
Md~Nayem Uddin, Kumar Shubham, Eduardo Blanco, Chitta Baral, and Gengyu Wang.
\newblock From recall to forgetting: Benchmarking long-term memory for
  personalized agents, 2026.
\newblock URL \url{https://arxiv.org/abs/2604.20006}.

\bibitem[Villalobos et~al.(2024)Villalobos, Ho, Sevilla, Besiroglu, Heim, and
  Hobbhahn]{villalobos2024running}
Pablo Villalobos, Anson Ho, Jaime Sevilla, Tamay Besiroglu, Lennart Heim, and
  Marius Hobbhahn.
\newblock Will we run out of data? limits of {LLM} scaling based on
  human-generated data, 2024.
\newblock URL \url{https://arxiv.org/abs/2211.04325}.

\bibitem[Wang et~al.(2026{\natexlab{a}})Wang, Duan, Wang, Zhang, and
  Wei]{wang2026streamprofilebench}
Sizhe Wang, Feiyu Duan, Juelin Wang, Liwen Zhang, and Zhongyu Wei.
\newblock {StreamProfileBench}: A benchmark for fine-grained user profile
  inference in real-world streaming scenarios, 2026{\natexlab{a}}.
\newblock URL \url{https://arxiv.org/abs/2605.25758}.

\bibitem[Wang et~al.(2025)Wang, Lu, Zhang, Huang, and Wang]{customerR1}
Ziyi Wang, Yuxuan Lu, Yimeng Zhang, Jing Huang, and Dakuo Wang.
\newblock Customer-r1: Personalized simulation of human behaviors via rl-based
  {LLM} agent in online shopping, 2025.
\newblock URL \url{https://arxiv.org/abs/2510.07230}.

\bibitem[Wang et~al.(2026{\natexlab{b}})Wang, Lu, Li, Amini, Sun, Bart, Lyu,
  Gesi, Wang, Huang, Su, Ehsan, Alikhani, Li, Chilton, and Wang]{wang2025opera}
Ziyi Wang, Yuxuan Lu, Wenbo Li, Amirali Amini, Bo~Sun, Yakov Bart, Weimin Lyu,
  Jiri Gesi, Tian Wang, Jing Huang, Yu~Su, Upol Ehsan, Malihe Alikhani, Toby
  Jia-Jun Li, Lydia Chilton, and Dakuo Wang.
\newblock {OPeRA}: A dataset of observation, persona, rationale, and action for
  evaluating {LLMs} on human online shopping behavior simulation,
  2026{\natexlab{b}}.
\newblock URL \url{https://arxiv.org/abs/2506.05606}.

\bibitem[Wu et~al.(2025{\natexlab{a}})Wu, Wang, Yu, Zhang, Chang, and
  Yu]{wu2024longmemeval}
Di~Wu, Hongwei Wang, Wenhao Yu, Yuwei Zhang, Kai-Wei Chang, and Dong Yu.
\newblock {LongMemEval}: Benchmarking chat assistants on long-term interactive
  memory, 2025{\natexlab{a}}.
\newblock URL \url{https://arxiv.org/abs/2410.10813}.

\bibitem[Wu et~al.(2025{\natexlab{b}})Wu, Ning, Liu, Lee, Wu, Wang, Prakash,
  O'Banion, Green, and Xie]{wu2024rlpf}
Jiaxing Wu, Lin Ning, Luyang Liu, Harrison Lee, Neo Wu, Chao Wang, Sushant
  Prakash, Shawn O'Banion, Bradley Green, and Jun Xie.
\newblock {RLPF}: Reinforcement learning from prediction feedback for user
  summarization with {LLMs}, 2025{\natexlab{b}}.
\newblock URL \url{https://arxiv.org/abs/2409.04421}.

\bibitem[Xu et~al.(2025)Xu, Chen, Jia, Zhou, Wu, Wang, Zhang, Zuo, Chen, Tang,
  Hu, and Zhu]{xu2025redrec}
Manjie Xu, Cheng Chen, Xin Jia, Jingyi Zhou, Yongji Wu, Zejian Wang, Chi Zhang,
  Kai Zuo, Yibo Chen, Xu~Tang, Yao Hu, and Yixin Zhu.
\newblock Cross-scenario unified modeling of user interests at billion scale,
  2025.
\newblock URL \url{https://arxiv.org/abs/2510.14788}.

\bibitem[Zhang et~al.(2026)Zhang, Tong, Fu, Zhao, Jiang, Jiang, and
  Yang]{lunar2026}
Jiahao Zhang, Yongzhi Tong, Zelin Fu, Pengde Zhao, Yanmei Jiang, Feng Jiang,
  and Min Yang.
\newblock {LUNAR}: Benchmarking personalized large language models on universal
  user behavior logs, 2026.
\newblock URL \url{https://arxiv.org/abs/2608.05246}.

\bibitem[Zhao et~al.(2026)Zhao, Hu, Song, Chen, Zhu, Xu, and
  Zheng]{llmauction2025}
Chujie Zhao, Qun Hu, Shiping Song, Dagui Chen, Han Zhu, Jian Xu, and Bo~Zheng.
\newblock {LLM-Auction}: Generative auction towards {LLM}-native advertising,
  2026.
\newblock URL \url{https://arxiv.org/abs/2512.10551}.

\bibitem[Zhao et~al.(2025{\natexlab{a}})Zhao, Hong, Liu, Hazarika, and
  Lin]{hou2025prefeval}
Siyan Zhao, Mingyi Hong, Yang Liu, Devamanyu Hazarika, and Kaixiang Lin.
\newblock Do {LLMs} recognize your preferences? evaluating personalized
  preference following in {LLMs}, 2025{\natexlab{a}}.
\newblock URL \url{https://arxiv.org/abs/2502.09597}.

\bibitem[Zhao et~al.(2025{\natexlab{b}})Zhao, Huang, Hu, Wang, Mao, Zhang,
  Zhang, Jiang, Wu, Ai, Wang, Zhou, and Chen]{zhao2024swift}
Yuze Zhao, Jintao Huang, Jinghan Hu, Xingjun Wang, Yunlin Mao, Daoze Zhang,
  Hong Zhang, Zeyinzi Jiang, Zhikai Wu, Baole Ai, Ang Wang, Wenmeng Zhou, and
  Yingda Chen.
\newblock Swift:a scalable lightweight infrastructure for fine-tuning,
  2025{\natexlab{b}}.
\newblock URL \url{https://arxiv.org/abs/2408.05517}.

\bibitem[Zheng et~al.(2025)Zheng, Liu, Li, Chen, Yu, Gao, Dang, Liu, Men, Yang,
  Zhou, and Lin]{gspo2025}
Chujie Zheng, Shixuan Liu, Mingze Li, Xiong-Hui Chen, Bowen Yu, Chang Gao, Kai
  Dang, Yuqiong Liu, Rui Men, An~Yang, Jingren Zhou, and Junyang Lin.
\newblock Group sequence policy optimization.
\newblock \emph{arXiv preprint arXiv:2507.18071}, 2025.

\bibitem[Zhou et~al.(2026)Zhou, Bao, Huang, Deng, Zhang, She, Cai, Ren, Ren,
  Luo, Wang, Hu, Zhang, Tang, Wang, Li, Wu, Luo, Wang, Hu, Wu, Liu, Zhang,
  Zhang, Chen, Wen, Ma, Song, Chu, Lian, Yang, Jiang, Cheng, Wang, Gai, Zheng,
  Wang, Huang, Mao, Gao, Yuan, Wang, Zhou, Su, Cheng, Ling, and
  Li]{zhou2026openonerec}
Guorui Zhou, Honghui Bao, Jiaming Huang, Jiaxin Deng, Jinghao Zhang, Junda She,
  Kuo Cai, Lejian Ren, Lu~Ren, Qiang Luo, Qianqian Wang, Qigen Hu, Rongzhou
  Zhang, Ruiming Tang, Shiyao Wang, Wuchao Li, Xiangyu Wu, Xinchen Luo, Xingmei
  Wang, Yifei Hu, Yunfan Wu, Zhanyu Liu, Zhiyang Zhang, Zixing Zhang, Bo~Chen,
  Bin Wen, Chaoyi Ma, Chengru Song, Chenglong Chu, Defu Lian, Fan Yang, Feng
  Jiang, Hongtao Cheng, Huanjie Wang, Kun Gai, Pengfei Zheng, Qiang Wang, Rui
  Huang, Siyang Mao, Tingting Gao, Wei Yuan, Yan Wang, Yang Zhou, Yi~Su, Zexuan
  Cheng, Zhixin Ling, and Ziming Li.
\newblock {OpenOneRec} technical report, 2026.
\newblock URL \url{https://arxiv.org/abs/2512.24762}.

\end{thebibliography}
\bibliographystyle{styles/colm2026_conference}

\appendix
\section{User Behavior Data}
\label{app:data-sources}

\subsection{Record Examples}
\label{app:records}
\par\medskip\noindent\fcolorbox{black!20}{black!2}{%
\begin{minipage}{\dimexpr\linewidth-2\fboxsep-2\fboxrule\relax}
\small\textbf{Synthetic behavior history}\par\smallskip
\texttt{[Day 1 09:10]} \textbf{Searches}: "lightweight backpack for a day hike"\par\smallskip
\texttt{[Day 2 19:30]} \textbf{Views}: Viewed "20 L daypack comparison".\par\smallskip
\texttt{[Day 3 08:20]} \textbf{Posts}: Posted "Planning a short day hike."\par\smallskip
\texttt{[Day 3 08:30]} \textbf{Follows}: Followed "Day-hiking tips" (fictional account).\par\smallskip
\texttt{[Day 4 18:00]} \textbf{Dislikes}: Hid a post about a 65 L expedition backpack.\par\smallskip
\medskip\textbf{Cutoff: Day 5 00:00. Hidden future follows.}\par\smallskip
\texttt{[Day 7 12:10]} \textbf{Purchases}: Purchased a 20 L daypack.\par\smallskip
\end{minipage}}\par\medskip

\section{Representative Task Templates}
\label{app:catalog}
Table~\ref{tab:representative-templates} gives one representative
template for each of the 21 nonempty behavior--question-type
combinations in Figure~\ref{fig:templates}. All questions use the
visible history $H_{<t}$. Predictive questions ask about behavior in
the hidden future $H_{\ge t}$; Objective questions additionally provide
candidates. Open questions analyze the visible history, while
Attribution questions disclose an observed future behavior and ask
for an explanation supported by earlier evidence. Bracketed text is
filled with task-specific materials during construction. Attribution
responses must identify evidence gaps and uncertainty, and must not
treat an observed association as proof of causation.

\begingroup
\small
\setlength{\tabcolsep}{4pt}
\begin{longtable}{@{}p{0.13\linewidth}p{0.20\linewidth}p{0.61\linewidth}@{}}
\caption{Representative task templates across six behavior categories
and four question types.}\label{tab:representative-templates}\\
\toprule
Behavior & Question type & Template \\
\midrule
\endfirsthead
\multicolumn{3}{@{}l}{Table~\thetable\ (continued)}\\
\toprule
Behavior & Question type & Template \\
\midrule
\endhead
\midrule
\multicolumn{3}{r@{}}{Continued on next page}\\
\endfoot
\bottomrule
\endlastfoot

Views & Objective &
Select the items in [candidate content] that the user will engage with
during [future window]. Engagement includes extended viewing, liking,
saving, commenting, or following; exposure or a click alone does not
qualify.\\
\addlinespace[4pt]
Views & Open-prediction &
Predict the topics and key entities of content the user will engage
with during [future window], without a candidate list.\\
\addlinespace[4pt]
Views & Open &
Summarize the user's content interests in the visible history. Support
each interest with observed behavior and state uncertainty when the
evidence is weak.\\
\addlinespace[4pt]
Views & Attribution &
Given [content the user engaged with], identify earlier topics,
content, or interactions that help explain this engagement.\\
\midrule

Purchases & Objective &
Select the products in [candidate products] that the user will purchase
during [future window]. Distinguish completed purchases from browsing,
clicking, or adding products to a cart.\\
\addlinespace[4pt]
Purchases & Open-prediction &
Predict which products the user will purchase during [future window],
without a candidate list. Keep the prediction at the product level.\\
\addlinespace[4pt]
Purchases & Open &
Summarize recurring product categories and product preferences
supported by the visible history. Cite supporting records and
distinguish browsing or comparison from completed purchases.\\
\addlinespace[4pt]
Purchases & Attribution &
Given [purchased products], explain how earlier product interactions,
searches, or comparisons relate to these purchases.\\
\midrule

Searches & Objective &
Select the queries in [candidate queries] that the user will search
for during [future window]. Do not treat content browsing or product
clicks as search queries.\\
\addlinespace[4pt]
Searches & Open-prediction &
Predict the queries the user will search for during [future window],
without a candidate list.\\
\addlinespace[4pt]
Searches & Open &
Answer [question about the user's history] using relevant records.
Distinguish supported conclusions from information that cannot be
determined from the available evidence.\\
\addlinespace[4pt]
Searches & Attribution &
Given [observed search queries], explain how earlier questions,
topics, or product-related behavior connect to these queries.\\
\midrule

Posts & Objective &
Select the topics in [candidate topics] that the user will publish
about during [future window]. Use the user's own earlier posts and
explicit creation-related evidence.\\
\addlinespace[4pt]
Posts & Open-prediction &
Predict the topics or content categories the user will publish about
during [future window], without candidates. Do not reconstruct the
future post's title or text.\\
\addlinespace[4pt]
Posts & Open &
Describe the user's writing style using only text authored by the user.
Support each style characteristic with examples from the visible history.\\
\addlinespace[4pt]
Posts & Attribution &
Given [observed publication topics], explain how the user's earlier
posts and changes in their content relate to these topics. Do not
treat other users' posts as the user's own work.\\
\midrule

Follows & Objective &
Select the accounts in [candidate accounts] that the user will follow
during [future window]. Distinguish following from viewing or liking
an account's content.\\
\addlinespace[4pt]
Follows & Open-prediction &
Predict the content themes of accounts the user will follow during
[future window]. Do not guess specific account identities.\\
\addlinespace[4pt]
Follows & Attribution &
Given [an account the user followed], explain how earlier exposure,
interactions, or expressed interests relate to this follow.\\
\midrule

Dislikes & Objective &
Select the items in [candidate content] that the user will explicitly
mark as uninteresting during [future window]. Lack of engagement alone
does not count as negative feedback.\\
\addlinespace[4pt]
Dislikes & Attribution &
Given [content the user hid], identify earlier negative feedback,
repeated exposure, or preference conflicts that could help explain
the response. Do not infer a definite reason without evidence.\\
\end{longtable}
\endgroup

\section{Composer and Scorer Prompts}
\label{app:prompts}

\begingroup
="[FandolSong-Regular.otf]" at 9pt

\newcommand{\velazh}[1]{{%
  \velapromptzh\XeTeXlinebreaklocale "zh"%
  \XeTeXlinebreakskip=0pt plus 1pt\relax
  #1}}
\newenvironment{velaprompt}{%
  \def\FrameCommand{\fcolorbox{black!20}{black!2}}%
  \MakeFramed{\advance\hsize-\width\FrameRestore}%
  \small\raggedright\setlength{\parskip}{4pt}%
}{\endMakeFramed}

\subsection{Composer System Prompt}
\label{app:composer-prompt}
\subsubsection*{English}
\begin{velaprompt}
You are the Task Composer. Given a selected template and the supplied
behavior records, construct zero or one valid question with its hidden
scoring specification. Return strictly valid JSON, without Markdown
or explanatory text outside the JSON object.

The request contains the frozen visible history
\texttt{user\_behaviors}, corresponding to $H_{<t}$, and a template
already selected by the program. Do not repeat template routing or
introduce a task outside the selected template. If the evidence does
not satisfy the template's requirements, return an empty candidate list.

Use the template's task definition, eligibility conditions, permitted
evidence, scoring requirements, and failure conditions to instantiate
the question. Preserve the solver-facing seed instruction, including
its task objective, analysis steps, output fields, and restrictions.
Do not copy template-editing headings or Composer-only instructions
into the question.

Place the task instruction first, followed by instance-specific
materials or a disclosed outcome, and then the output requirements.
If the seed instruction already specifies the output format, integrate
any additional instance-specific constraints into that format rather
than adding a second output specification. Do not duplicate the full
visible history or the final model input inside the candidate.

For Open questions, use only the visible history. Construct the question
and a task-specific rubric without producing a unique reference answer,
\texttt{gold}, or \texttt{gold\_items}. The answer must be evaluated
against the question requirements and visible evidence, not against
a summary written by the Composer.

For Attribution questions, the request additionally supplies observed
future behavior in \texttt{future\_user\_behaviors}, corresponding to
$H_{\ge t}$. Use the relevant observed behavior as the disclosed outcome
to be explained. Include the outcome information required by the
template in the question. The explanation must be supported by the
visible history; do not use undisclosed future information as evidence.

When a disclosed outcome requires a content record, include its relevant
title, text, and available image or video descriptions. When the template
requires a complete exposure list, preserve its order and distinguish
recorded engagement from exposure alone. Skip the candidate if required
materials are missing.

Do not invent entities, events, conflicts, or missing context. Do not
upgrade weaker behavior into a stronger outcome, such as treating a
click as a purchase. Keep uncertainty and alternative explanations when
the evidence does not support a unique conclusion. Do not reveal hidden
scoring answers, rubric weights, or scoring anchors in the question.

Derive the rubric from the complete template and the evidence available
for this instance. Use non-overlapping, independently assessable
criteria, typically four to seven. Each criterion must identify a
concrete requirement and define \texttt{full\_credit},
\texttt{partial\_credit}, and \texttt{zero\_credit}. Assign equal
weight $1/m$ to each of the $m$ criteria.

Distinguish factual accuracy from coverage where relevant. Accuracy
checks entities, behavior, quantities, timing, and evidence references.
Coverage checks whether the response addresses the requested targets,
fields, evidence, counterevidence, and conclusions. Avoid counting the
same requirement under multiple criteria.

Every assessed requirement must be stated clearly in the question.
If the rubric checks temporal relationships, counterevidence, or
uncertainty, the question must request that analysis. Do not introduce
hidden requirements or artificial difficulty unsupported by the records.

For each scoring subtask, provide a stable, unique
\texttt{subtask\_id}, \texttt{metric="rubric"},
\texttt{rubric\_dimensions}, \texttt{hard\_fail\_conditions}, and
\texttt{evidence\_scope}. For Open questions, also include
\texttt{required\_coverage} and \texttt{evidence\_anchors} containing
record identifiers and minimal supporting facts. Evidence anchors are
references for verification, not a model-written reference answer.

Return \texttt{\{"candidates": [...]\}} using the supplied schema.
Include the instantiated question and hidden scoring specification,
with no additional candidates or unsupported information.
\end{velaprompt}

\subsubsection*{Chinese}
\begin{velaprompt}
\velazh{%
你是 Task Composer。根据已经选定的模板和提供的行为记录，构建零个或一个合格问题及其隐藏评分合同。
只输出严格 JSON，不输出 Markdown 或 JSON 对象以外的说明。

请求包含冻结的可见历史 \texttt{user\_behaviors}，对应 $H_{<t}$，以及已经由程序选定的模板。
不要重复判断模板路由条件，不得生成所选模板范围之外的任务。如果证据不满足模板要求，返回空候选列表。

综合模板中的任务定义、出题条件、可用信息、评分要求和不合格条件实例化问题。
保留面向答题者的种子提示词，包括任务目标、分析步骤、输出字段和业务限制。
不要把模板编辑标题或仅供 Composer 使用的说明复制进题面。

题面先写任务指令，再写本实例的供给材料或已披露结果，最后写输出要求。
如果种子提示词已经规定输出格式，将实例专属的新限制合并到该格式中，不要追加第二份输出合同。
不要在候选中重复复制完整可见历史或最终模型输入。

普通开放题只能使用可见历史。生成具体问题和任务专属 rubric，不生成唯一参考答案、
\texttt{gold} 或 \texttt{gold\_items}。评分应依据题面要求和可见证据，
不能把 Composer 自己写的摘要当作标准答案。

归因题还会收到 \texttt{future\_user\_behaviors}，对应隐藏未来 $H_{\ge t}$。
从中选取相关的真实行为作为题面披露的解释对象，并公开模板要求的结果信息。
解释必须由可见历史支持，不得使用题面未披露的未来信息补充证据。

已披露结果需要对应内容记录时，应提供相关标题、正文以及可用的图片和视频描述。
模板要求完整曝光列表时，保留原排序，并区分已记录的互动和单纯曝光。
缺少必要材料时跳过该候选。

不得编造实体、事件、冲突或缺失背景，不得把弱行为升级为更强的结果，例如把点击写成购买。
证据不足以支持唯一结论时，应保留不确定性和替代解释。
不得在题面中泄露隐藏评分答案、rubric 权重或评分锚点。

根据完整模板和当前实例的证据生成 rubric。评分维度应互不重复、能够独立判断，通常为四至七个。
每个维度对应一个具体要求，并定义 \texttt{full\_credit}、\texttt{partial\_credit}
和 \texttt{zero\_credit}。对 $m$ 个维度采用相同权重 $1/m$。

需要时区分信息准确性和信息完整性。准确性检查实体、行为、数值、时间和证据引用；
完整性检查题面要求的目标、字段、证据、反向信号和结论是否全部覆盖。
同一要求不得在多个维度重复计分。

每个评分要求都必须在题面中明确表达。如果 rubric 检查时间关系、反向信号或不确定性，
题面必须要求相应分析。不得设置隐藏考点，也不得虚构记录不支持的难点。

每个评分子题包含稳定且唯一的 \texttt{subtask\_id}、\texttt{metric="rubric"}、
\texttt{rubric\_dimensions}、\texttt{hard\_fail\_conditions} 和 \texttt{evidence\_scope}。
普通开放题还需包含 \texttt{required\_coverage} 和 \texttt{evidence\_anchors}；
后者只保存记录编号与最小事实摘要，用于核验证据，不能作为模型生成的标准答案。

按请求提供的 schema 返回 \texttt{\{"candidates": [...]\}}，
包含具体问题和隐藏评分合同，不添加额外候选或缺乏证据的信息。
}
\end{velaprompt}

\subsection{Scorer System Prompt}
\label{app:scorer-prompt}
\subsubsection*{English}
\begin{velaprompt}
You are the Scorer. Evaluate one response from the evaluated LLM at a
time. The input contains the task, permitted evidence, hidden scoring
specification, response protocol, and an \texttt{output\_contract}
for your scoring result. Do not rewrite the task or introduce new
scoring requirements. Return strictly valid JSON.

Identify each subquestion's response and match its scoring specification
to its response protocol by \texttt{subtask\_id}, not by array position.
Different subquestions may require different formats. Apply each format
only to its own subquestion; a response containing multiple formats must
not be rejected merely because it is not one JSON object.

For Objective questions, extract the evaluated LLM's final selected
candidate identifiers conservatively and return them as
\texttt{selected\_ids}. Do not add selections, correct the answer, or
infer a choice that was not made. Follow the question's permitted answer
format when interpreting identifiers. Ground-truth comparison and set-F1
are computed by the scoring program; the extraction step does not
require access to hidden ground truth.

For an Open-prediction question, you receive the ground-truth answer
items derived from the hidden future and the evaluated LLM's free-form
response. Extract the predicted items from the response according to
the question's requested target type and level of detail. Do not infer
additional predictions or use external knowledge to repair the answer.

Identify semantically equivalent prediction--ground-truth pairs.
The behavior type and target type must agree. Evaluate the relevant
topic, entity, product, brand, or category within the same pair; do not
combine attributes from different ground-truth items to create a match.
Broad topical similarity or a partially correct entity does not by
itself establish semantic equivalence.

Use one-to-one matching: each predicted item can match at most one
ground-truth item, and each ground-truth item can be matched at most
once. Resolve duplicate expressions consistently so that repeated
predictions cannot receive repeated credit.

Map each semantically equivalent prediction to its corresponding
ground-truth item. Retain unmatched predictions in the predicted set
so that incorrect predictions affect precision, and retain the complete
ground-truth set so that omissions affect recall. Return the extracted
predictions and semantic correspondences as required by the output
contract. The scoring program then computes ordinary set-F1 from the
normalized predicted set and ground-truth set. Semantic matching does
not assign fractional match scores or add separate evidence or format
bonuses.

For Open and Attribution questions, evaluate each criterion against its
\texttt{full\_credit}, \texttt{partial\_credit}, and
\texttt{zero\_credit} descriptions and assign a score in $[0,1]$.
The question score is the equal-weight mean of the criterion scores.
Do not assign an overall impression score first and construct reasons
afterward. If an explicit hard-failure condition is met, return zero
for that question.

Use only evidence allowed by \texttt{evidence\_scope}. For Open
questions, verify facts and coverage against the visible history and
the question requirements. There is no unique reference answer.
Use \texttt{evidence\_anchors} to locate supporting records, not as
text that the response must reproduce. Disregard any mistakenly supplied
hidden future or reference answer when scoring an Open question.

For Attribution questions, the disclosed outcome is the object of
explanation. Judge whether the explanation is supported by the visible
history and addresses any requested counterevidence, temporal
relationships, alternatives, and uncertainty. Do not treat an observed
association as proof of causation.

Check that cited record identifiers exist and that quoted evidence
supports the claims. Do not accept fabricated facts, unsupported
identity or health inferences, or stronger intentions inferred solely
from weak behavior. For memory or profile tasks, apply only the
dimensions specified by that task's rubric. Do not penalize reasonable
differences in wording, ordering, or organization, or impose additional
style and length requirements.

Follow the supplied \texttt{output\_contract}. Its metadata defines the
required structure and must not be copied as answer fields. For a
composite task, return \texttt{subtask\_scores} with each requested
\texttt{subtask\_id} exactly once. For a top-level rubric request,
return the requested top-level score instead. Include brief reasons
only where permitted. The scoring program aggregates question scores
into the task score. If validation feedback is supplied, correct the
specified output error rather than repeating an invalid JSON result.
\end{velaprompt}

\subsubsection*{Chinese}
\begin{velaprompt}
\velazh{%
你是 Scorer。每次只评价被评测 LLM 的一份回答。输入包含任务、允许使用的证据、
隐藏评分合同、答题协议，以及规定评分结果格式的 \texttt{output\_contract}。
不得重新出题或增加评分要求，只输出严格 JSON。

识别每个子题的回答，并按 \texttt{subtask\_id} 将评分合同与答题协议配对，
不要按数组位置推测对应关系。不同子题可以使用不同格式；
每种协议只约束对应子题，不能因为整份回答包含多种格式、不是单个 JSON 对象就判为无效。

对于 Objective 题，保守抽取被评测 LLM 最终选择的候选编号，返回 \texttt{selected\_ids}。
不得补选、纠错，或推断回答没有明确作出的选择。按题面允许的格式解释候选编号。
由评分程序与隐藏 ground truth 比较并计算 set-F1；抽取步骤不需要访问隐藏答案。

对于 Open-prediction 题，输入包含从隐藏未来提取的 ground truth，以及被评测 LLM 的自由形式回答。
按题面指定的目标类型与粒度抽取预测项，不得额外推断预测内容，也不得用外部知识修补回答。

识别语义等价的预测项与 ground truth 配对。行为类型和目标类型必须一致。
主题、实体、商品、品牌或类目必须在同一对预测项与 ground truth 内判断，
不能把不同 ground truth 的属性拼接成一次命中。
仅有宽泛主题相似或部分实体正确，不足以构成语义等价。

采用一对一匹配：每个预测项最多匹配一个 ground truth，每个 ground truth 最多被匹配一次。
一致处理重复表述，避免重复预测获得多次得分。

将语义等价的预测项映射为对应的 ground-truth 项。
未匹配的预测项仍保留在预测集合中，使错误预测影响 precision；
保留完整 ground-truth 集合，使遗漏影响 recall。
按输出合同返回抽取的预测项与语义对应关系，由评分程序根据归一化预测集合和
ground-truth 集合计算普通 set-F1。语义匹配不给部分匹配分，也不额外混入证据分或格式分。

对于 Open 和 Attribution 题，逐维对照 \texttt{full\_credit}、\texttt{partial\_credit}
和 \texttt{zero\_credit}，给出 $[0,1]$ 范围内的分数。
题目得分为各维度分数的等权平均。不得先凭整体印象打分，再反推理由。
命中明确规定的硬失败条件时，该题记零分。

只能使用 \texttt{evidence\_scope} 允许的证据。
普通开放题根据可见历史和题面要求核对事实与覆盖情况，没有唯一参考答案。
\texttt{evidence\_anchors} 只用于定位证据，不能要求回答复述其文字。
如果输入错误地包含隐藏未来或参考答案，不得据此评价普通开放题。

归因题以题面已经披露的结果为解释对象。检查解释是否得到可见历史支持，
并按题面要求检查反向信号、时间关系、替代解释和不确定性。
不得将观测到的相关关系当作因果证明。

核对引用的记录编号是否存在，以及引用内容是否支持结论。
不得接受编造事实、缺乏依据的身份或健康推断，或仅凭弱行为就断言更强意图。
对于 memory 或 profile 任务，只使用该任务 rubric 指定的维度。
不得因合理的措辞、排序或组织差异扣分，也不得擅自增加文风和篇幅要求。

遵守 \texttt{output\_contract}。合同中的元数据用于规定结构，不是需要复制的输出字段。
复合任务返回 \texttt{subtask\_scores}，每个要求的 \texttt{subtask\_id} 恰好出现一次；
顶层 rubric 请求则返回规定的顶层分数。只有合同允许时才附简短理由。
评分程序将各题得分汇总为任务得分。收到校验错误反馈时，修正指出的错误，不要重复无效 JSON。
}
\end{velaprompt}

\endgroup

\clearpage
\section{Benchmark Examples}
\label{app:benches}

\begin{center}
\small
\setlength{\tabcolsep}{5pt}
\begin{tabular}{@{}>{\raggedright\arraybackslash}p{0.24\linewidth}>{\raggedright\arraybackslash}p{0.24\linewidth}>{\raggedright\arraybackslash}p{0.45\linewidth}@{}}
\toprule
Benchmark & Evaluation set & Reported metric \\
\midrule
PurchaseIntent & 500 users; 1,137 labels &
recall@20, with all ground-truth purchases in the denominator \\
ProfileSummary & 230 examples &
LLM-based score: profile (0.3), core interests (0.5), inferred interests (0.2) \\
PersonaBench & 3,690 questions; 32 subsets &
Macro-F1 across subsets \\
AdRecommendation & 996 examples &
LLM-based rubric score for query rewrites \\
\bottomrule
\end{tabular}
\end{center}

\begingroup
="[FandolSong-Regular.otf]" at 9pt

\newcommand{\benchzh}[1]{{\velabenchzh\XeTeXlinebreaklocale "zh"%
  \XeTeXlinebreakskip=0pt plus 1pt\relax #1}}
\newenvironment{velabenchexample}[1]{%
  \par\medskip\noindent\begin{minipage}{\linewidth}%
  \def\FrameCommand{\fcolorbox{black!20}{black!2}}%
  \MakeFramed{\advance\hsize-\width\FrameRestore}%
  \small\raggedright\setlength{\parskip}{4pt}\textbf{#1}\par
}{\endMakeFramed\end{minipage}\par}

\begin{velabenchexample}{PurchaseIntent}
\benchzh{汉堡小食三件套随心配 【到店用】特惠团购电子优惠券全国通用 | 兑换券 | 19天前 | instant\_buy×1 click×1 goods\_view×1 笔记点商品×1}\par
\benchzh{[\ldots]}\par
\benchzh{麻酱拌冷面速食夏季拌面 | 粮油调味/速食/干货/烘焙 | 14天前 | click×2 进店×1 goods\_view×2 [\ldots]}\par
\textbf{Question}\par
\benchzh{请预测用户接下来最可能购买哪些候选商品 [\ldots]}\par
\textbf{Ground truth}: \benchzh{麻酱拌冷面速食夏季拌面}
\end{velabenchexample}

\begin{velabenchexample}{ProfileSummary}
\textbf{Search history}\par
\benchzh{搜:新手买拼豆买几色的好 27天前}\par
\benchzh{搜:拼豆用豆针还是镊子 14天前}\par
\benchzh{搜:拼豆电熨斗推荐 3天前}\par
\textbf{Task}\par
\benchzh{根据用户行为数据输出结构化画像 + 兴趣总结 + 推理兴趣。}\par
\textbf{Rubric reference}\par
\benchzh{兴趣实体：拼豆手工DIY(材料套装/图纸/工具)}
\end{velabenchexample}

\begin{velabenchexample}{PersonaBench}
\textbf{Shopping history}\par
\benchzh{加购物车：[\ldots]泰式无皮虾鲜活味道开背生呛虾生腌虾即食零食}\par
\textbf{Question}\par
\benchzh{请根据以下用户行为数据和前期购买记录，预测该用户下一单最可能购买哪个品类的商品：}\par
\benchzh{A. 服饰鞋包 \quad B. 食品生鲜 \quad C. 美妆个护 \quad D. 母婴亲子 \quad E. 家居日用}\par
\textbf{Ground truth}: \texttt{B}
\end{velabenchexample}

\begin{velabenchexample}{AdRecommendation}
\textbf{Recent searches}\par
\benchzh{豆腐做法；家常菜清淡做法；高笋炒肉；家常荤菜；素菜家常菜；家常菜菜谱图片}\par
\textbf{Task}\par
\benchzh{需要综合分析后输出面向广告召回的高商业价值改写词。}\par
\textbf{Rubric criteria}\par
\benchzh{cate: 美食}\par
\benchzh{entity: 家常菜、豆腐、蔬菜、减脂、清淡、菜谱、高笋}\par
\benchzh{diverge: 厨房小家电、低脂调味品、健康食材、烹饪工具、生鲜食材电商、地方特色调味酱料、快手半成品菜}
\end{velabenchexample}
\endgroup

\clearpage
\section{RL Configuration}
\label{app:rl-config}
We conduct four RL experiments, using Qwen3.6-35B-A3B and Vela-35B-A3B
as initializations in each of the VelaBench and ProfileSummary domains.
All four experiments use the configuration in Table~\ref{tab:rl-config}.
For the VelaBench experiment, we randomly sample 2{,}000 tasks from the
training set. For the ProfileSummary experiment, we use 2{,}330 training
examples that require a structured user profile, core interests, and
inferred interests. Within each experiment, both model initializations
use the same RL training data and configuration.
RL training and evaluation data have no overlapping samples or users.
Vela-35B-A3B is initialized from the SFT checkpoint trained on 18.8K
Vela-Instruct examples, as in Section~\ref{sec:rl}.

\paragraph{Training configuration.}
We use Group Sequence Policy Optimization (GSPO; \citealp{gspo2025})
with 16 responses per prompt and no learned critic.
Training uses Megatron-LM, while SGLang generates rollouts.
Of the 48 GPUs, 32 are allocated to policy training and 16 to
rollout inference. Table~\ref{tab:rl-config} lists the configuration.

\begingroup
\small
\setlength{\tabcolsep}{6pt}
\begin{longtable}{@{}>{\raggedright\arraybackslash}p{0.49\linewidth}>{\raggedright\arraybackslash}p{0.46\linewidth}@{}}
\caption{Shared reinforcement learning configuration for all four experiments.}
\label{tab:rl-config}\\
\toprule
Parameter & Value \\
\midrule
\endfirsthead
\multicolumn{2}{@{}l}{Table~\thetable\ (continued)}\\
\toprule
Parameter & Value \\
\midrule
\endhead
\bottomrule
\endfoot
\multicolumn{2}{@{}l}{\textit{Optimization}}\\
Optimizer & Adam \\
Learning rate & 0.000001, constant; no warmup \\
Adam beta coefficients & $(0.9,0.98)$ \\
Weight decay / gradient clipping norm & $0.1$ / $1.0$ \\
KL loss / coefficient & Low-variance KL / $0.001$ \\
Clipping $\epsilon$ / upper clipping $\epsilon$ & $0.2$ / $0.28$ \\
Entropy coefficient & $0$ \\
Advantage normalization & Disabled \\
Training precision / gradient reduction & BF16 / FP32 \\
Optimizer CPU offload & Enabled \\
Precision-aware optimizer & Enabled \\
\midrule
\multicolumn{2}{@{}l}{\textit{Rollouts and training budget}}\\
Prompts per rollout batch & 24 \\
Responses per prompt & 16 \\
Global batch size & 384 responses \\
Configured rollout iterations & 400 \\
Checkpoint saving interval & 50 rollout steps \\
Sampling temperature & $1.0$ \\
Maximum response length & 1{,}536 tokens \\
Rollout thinking mode & Disabled \\
Dynamic batching / token budget per GPU & Enabled / 2{,}048 tokens \\
SGLang memory fraction / concurrency & $0.85$ / 24 \\
Over-sampling batch size & 24 \\
\midrule
\multicolumn{2}{@{}l}{\textit{Parallelism}}\\
Tensor / pipeline / expert / context parallelism & $2$ / $1$ / $8$ / $2$ \\
Expert tensor parallelism & 1 \\
Sequence parallelism & Enabled \\
Activation recomputation & Full, uniform; 2 layers \\
Tensor parallelism per rollout engine & 2 \\
\midrule
\multicolumn{2}{@{}l}{\textit{Reward scoring}}\\
Reward scorer & Qwen3.8-2.4T-A95B \\
Scorer thinking level & Low \\
Scorer maximum output length & 16{,}000 tokens \\
Scorer concurrency & 128 \\
\end{longtable}
\endgroup

\paragraph{Reward computation.}
For ProfileSummary, the reward scorer evaluates the generated user
profile, core interests, and inferred interests using task-specific
rubrics.

\end{document}